\documentclass[review]{elsarticle}
\makeatletter
\def\ps@pprintTitle{%
 \let\@oddhead\@empty
 \let\@evenhead\@empty
 \def\@oddfoot{\centerline{\thepage}}%
 \let\@evenfoot\@oddfoot}
\makeatother
\usepackage{hyperref}
\usepackage{graphicx}
\usepackage{amsmath,amssymb} 
\usepackage{subfig}
\usepackage{picins}

\newcommand{\ignore}[1]{}
\begin{document}
\begin{frontmatter}
\title{Blur-Countering Keypoint Detection via Eigenvalue Asymmetry}
%
%



\author[mymainaddress]{Chao Zhang}

\author[mysecondaryaddress]{Xuequan Lu}

\author[mythirdaddress]{Takuya Akashi}

\address[mymainaddress]{3-9-1 Bunkyo,University of Fukui, Japan}
\address[mysecondaryaddress]{50 Nanyang Avenue, Nanyang Technological University, Singapore}
\address[mythirdaddress]{4-3-5 Ueda, Iwate University, Japan}
\begin{abstract}
Well-known corner or local extrema feature based detectors such as FAST and DoG have achieved noticeable successes. However, detecting keypoints in the presence of blur has remained to be an unresolved issue. As a matter of fact, various kinds of blur (e.g., motion blur, out-of-focus and space-variant) remarkably increase challenges for keypoint detection. As a result, those methods have limited performance. To settle this issue, we propose a blur-countering method for detecting valid keypoints for various types and degrees of blurred images. Specifically, we first present a distance metric for derivative distributions, which preserves the distinctiveness of patch pairs well under blur. We then model the asymmetry by utilizing the difference of squared eigenvalues based on the distance metric. To make it scale-robust, we also extend it to scale space. The proposed detector is efficient as the main computational cost is the square of derivatives at each pixel. Extensive visual and quantitative results show that our method outperforms current approaches under different types and degrees of blur. Without any parallelization, our implementation\footnote{We will make our code publicly available upon the acceptance.} achieves real-time performance for low-resolution images (e.g., $320\times240$ pixel).
\end{abstract}

\end{frontmatter}


\section{Introduction}
Keypoint detection, a fundamental technique in computer vision, has gained extensive attention in recent decades. It plays an important role in various applications such as image retrieval \cite{zhou2016scalable,zheng2017sift}, image stitching \cite{li2018quasi,szeliski2006image}, object recognition \cite{soltanpour2017survey,mikolajczyk2005local} and so on. It typically requires finding pixels or blobs which are supposed to be invariant against either photometric or geometric variations. Most of the existing methods attempt to improve the robustness against photometric variations from two different aspects: \textit{methods with utilization of sharp features and data-driven approaches}. Nevertheless, these techniques are limited in the presence of image blur. Regarding the former class, the intersection of two edges may be smoothed out by image blur and lose the distinctiveness despite the fact that corner-feature based detectors such as Harris \cite{harris1988combined} can be robust to illumination changes. A more popular way is to find local extremes over scale space generated by different sizes of Gaussians \cite{bay2006surf,lindeberg1994scale,lowe2004distinctive}. However, in the case of motion blur, when the illuminance changes are integrated along a specified direction over time, the positions of local extremes will not be guaranteed the same, because the directional average may change the local distribution of intensities. For the latter category, one can collect training sets which will indirectly determine the type of feature detectors \cite{lepetit2006keypoint,ozuysal2010fast,rosten2010faster,altwaijry2016learning} by casting the keypoint detection task as a classification problem. Large scale of patches with good keypoints can be collected and annotated in the case of unblurred images. However, the definition of ``good keypoints'' involving both unblurred and blurred patches would become too ambiguous to supervise.

\begin{figure}[t]
	\centering
\subfloat[Fast-Hessian]{
\includegraphics[width=0.4\linewidth]{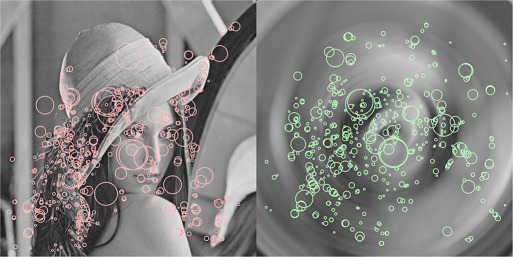}
 }
\subfloat[Fast-Hessian]{
\includegraphics[width=0.4\linewidth]{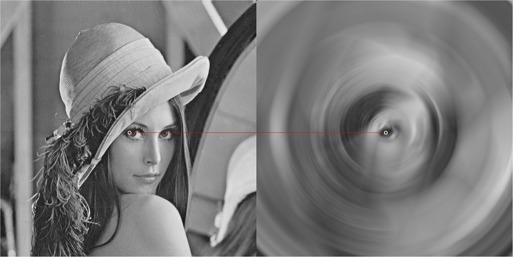}
}

\subfloat[Our method]{
\includegraphics[width=0.4\linewidth]{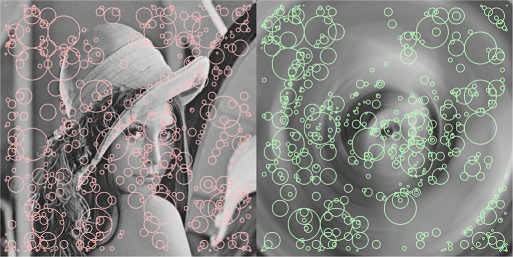}
}
\subfloat[Our method]{
\includegraphics[width=0.4\linewidth]{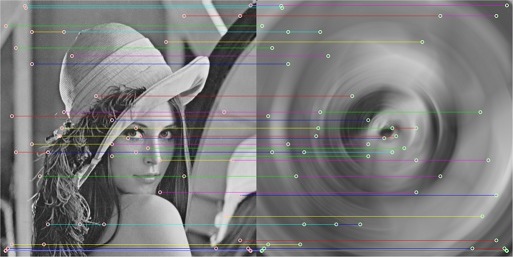}
}
\caption{Keypoint detection results under drastic rotational blur (60 degree, with its center located at the image center). Each pair of corresponding points is connected by a line only if their coordinates from two images are exactly the same. To only focus on analyzing the performance of keypoint detectors, no feature descriptor is combined for feature matching throughout this paper. \textbf{(a)} Top-500 keypoints detected by Fast-Hessian from SURF \cite{bay2006surf} in both blurred/unblurred images are shown. Blobs detected by Fast-Hessian tend to be aggregated in regions with sharp features. \textbf{(b)} Corresponding points based on the detected keypoints by Fast-Hessian. Only the points detected in the center without influence of blur are matched. \textbf{(c)} Top-500 keypoints detected by our method. Our detected keypoints have a more uniform distribution and higher correspondences. \textbf{(d)} 51 pairs of corresponding points are matched from top-500 keypoints. In this work, the parameters of our detector are fixed for all experiments. }
\label{fig:visualcomparison}
\end{figure}

Image blur is pervasive in videos and images, mainly because of fast motion during exposure time and imperfect auto-focus systems. It destroys sharp features like edges, corners and local extremes. As shown in Fig. \ref{fig:visualcomparison}(a) and (b), with the influence brought by rotational blur, the intensities change drastically and the conventional method falls in failure. Detecting corresponding keypoints with image blur, which has been sparsely treated so far, suffers from the following challenges: (1) runtime performance. It is of critical
importance because keypoint detection is usually used as the first step in many real-time applications such as SLAM \cite{williams2007real}. (2) Image degradation. Regardless of loosing sharp features mentioned above, strong motion blur can possibly introduce additional features, as a result of ``stretching'' pixels along motion direction. For instance, the rotational blur in Fig. \ref{fig:visualcomparison} produces edges in concentric circles, leading to unwanted detections. (3) Various types and degrees of blur. Motion in practice can be space-variant and complex, difficult to be inferred and separated. One natural idea for keypoint detection with blur is to adopt blind image deblurring algorithms \cite{pan2016blind,nah2017deep} as preprocessing. However, the iterative deblurring is often time-consuming. Also, it may classify strong artifacts as true features, which will thus decrease the performance of detectors.

The main technical contributions of this paper are threefold. First, we demonstrate that the distance between two different derivative distributions of image patches can well keep its distinctiveness even in the presence of image blur. Second, to decrease the cost of dense distance calculation, we use the difference of the sum of squared eigenvalues to measure the distance between derivative distributions. Since the sum of eigenvalues can be calculated from squared derivatives only, this scheme can be efficient. Eventually, instead of scoring likelihood of geometric features or discontinuity of intensities, the degree of asymmetry is calculated from neighbor patches' derivative distributions in different radial directions and used for ranking the keypoint candidates. In this work, the proposed method or detector is named as eigenvalue asymmetry (EAS) for simplicity and convenience.

\section{Related Work}
We only review previous techniques which are mostly related to our work. Interested readers are referred to survey papers \cite{tuytelaars2008local,krig2016interest} for a more comprehensive review of state-of-the-art keypoint detectors.

Symmetry has been exploited for keypoint detection in several works. Loy et al. \cite{loy2003fast} developed a fast radial symmetry transform to detect keypoints. With a determined circular range, gradients at two points (i.e., positively-affected point and negatively-affected point) are examined to calculate a symmetry contribution. By summing the contributions over all the ranges considered, the degree of symmetry can be calculated. However, the pixels with small gradient magnitudes are sensitive to noise and would provide unreliable orientations when determining the two points mentioned above. Besides, the detection results tend to be located at edges. Tong et al. \cite{tong2017blur} proposed a keypoint detector based on blur-invariant moments \cite{flusser1998degraded}. An image region is divided into two portions according to a certain symmetrical axis. Differences of intensity, center, skewness explained by moments are incorporated to evaluate the degree of symmetry. In their implementation, five types of fixed circular filters need to be convolved, which is time-consuming and rotation-variant. Hauagge et al. \cite{Hauagge2012} introduced a local feature detector for matching with cross-domain images, which is robust with large appearance changes. Both bilateral and rotational symmetries are considered to build a distance measure with either raw intensities or dense gradient histograms. Still, it is computationally expensive to score the degree of symmetry by computing SIFT or HOG at each pixel. In this work, we propose to measure the degree of asymmetry by averaging distances between derivative distributions radially to avoid the estimation of hypothetical axis of asymmetry.

Low self-similarity \cite{moravec1980obstacle,smith1997susan,trajkovic1998fast} is a classical idea which utilizes the fact that the changes of intensity should be high in all directions around a highly distinctive corner. Following this, the basic scheme is to design a comer response function which evaluates the ``cornerness'' by calculating the changes of intensity over pixels/patches. It has a close relationship with the idea of symmetry/asymmetry, as a region with high degree of asymmetry generally appears to be involved in low self-similarity. It should be noted that self-similarity does not take spatial information into account, yet the measurement of symmetry/asymmetry structure considers the spatial information. 

Eigenvalues have been used for modeling keypoint detectors. Harris et al. \cite{harris1988combined} treat an area as a corner when both eigenvalues of the covariance matrix are positively large. Shi et al. \cite{shi1994good} claimed that corresponding area can be regarded as a good candidate of features when the smaller eigenvalue is sufficiently large to meet the noise criterion. In this work, we employ eigenvalues to measure the distance between derivative distributions, which in essence estimates the asymmetry and keypoint likelihood.

\section{Proposed method}

In this section, we attempt to design a keypoint metric which maintains the order of keypoint scores measured throughout the image consistently in blurred and unblurred images. In the ideal case, the $TopN$ keypoints from a unblurred image and its blurred version should be the same and correspond to each other.

\subsection{Distance between Derivative Distributions}
\begin{figure}[t]
	\centering
\subfloat[Unblurred patches]{
\includegraphics[width=0.95\linewidth]{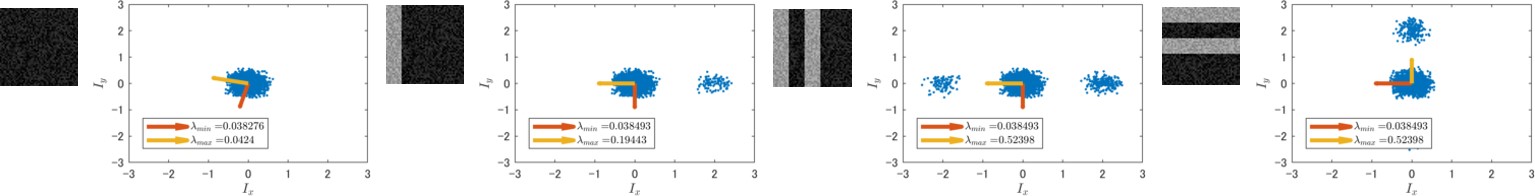}
 }
 
\subfloat[Bluured patches]{
\includegraphics[width=0.95\linewidth]{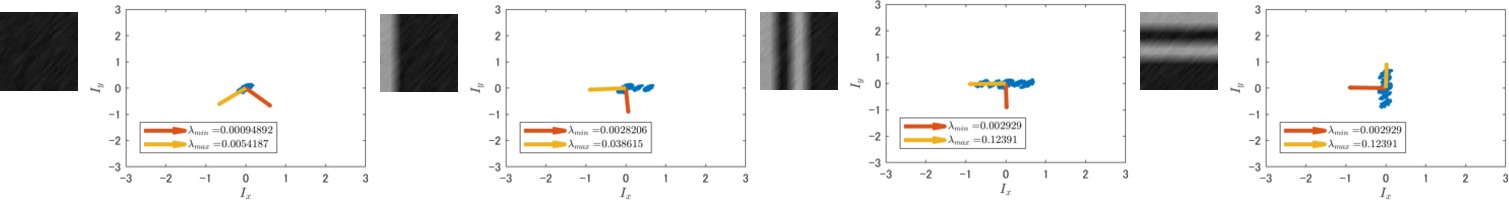}
}
\caption{(a) Four different types of patches and their derivative distributions are shown. (b) Derivative distributions of the corresponding blurred patches to the above. Here, we set the length of motion $l=10$ and angle of motion $\theta=\pi/4$. }
\label{fig:patchtest}
\end{figure}

Instead of estimating similarity between two patches with raw intensities or gradient histograms, we propose to measure the similarity between two different derivative distributions. Derivative distribution refers to the distribution of image derivatives, with horizontal axis representing the horizontal derivative, and vertical axis representing the vertical derivative, which is shown in \mbox{Fig. \ref{fig:patchtest}}. This can be treated as a problem of distance measurement between two point sets $p$ and $q$, with each element from the two sets consisting of derivative $I_x$ and $I_y$. Each similarity calculated from two point sets contribute to the final score of asymmetry, which will be introduced later. Various methods offer such distance measurement such as BBS \cite{dekel2015best}, EMD \cite{rubner2000earth}, etc. However, they all require computing point-to-point distances, which could be computationally expensive in our task. In the case of BBS, assuming $\vert p \vert =s$ and $\vert q \vert =s$, the number of tests is $d$, then the computational complexity for scoring asymmetry with image size $\vert I \vert=L$ is $O(dLs^2)$. Since the eigenvalues measure the variances along each eigenvector of the point sets, the shapes of $p$ and $q$ can be modeled by two ellipses with the eigenvalues as the lengths of semi-axes. We then define the distance between $p$ and $q$ as
\begin{equation}
\mathrm{d}(p,q) = \vert (\lambda_{max}^p+\lambda_{min}^p)-(\lambda_{max}^q+\lambda_{min}^q) \vert,
\label{Eq:dpq}
\end{equation}
where $\lambda_{max}$ represents the length of the semi-major axis while $\lambda_{min}$ represents the length of the semi-minor axis. They are calculated from the covariance matrix of $p$ and $q$, respectively.

To investigate the usefulness of this distance metric, we first analyze the sum of the semi-major and semi-minor axes. The perimeter of an ellipse parameterized by $\lambda_{max}$ and $\lambda_{min}$ can be represented by $(\lambda_{max}+\lambda_{min})\pi \sum_{z=1}^{\infty} 
\mathrm{C}(0.5,n)h^z$, where $h=(\lambda_{max}-\lambda_{min})/(\lambda_{max}+\lambda_{min})$, $h\in[0,1]$ and $\mathrm{C}$ is the combination function. Obviously, $\lambda_{max}+\lambda_{min}$ dominates the perimeter, which indicate that Eq. (\ref{Eq:dpq}) calculates the distance mostly based on the difference of perimeters and ignores the uniformity of data distribution. The distance metric is translation- and rotation-invariant: applying translation and rotation to the data points does not change the distance. This characteristic can help to endure the changes of distribution's shape at some extent caused by noise. The reason is that with the perimeter fixed, the shape of the ellipse can vary within a tolerance range.\ignore{On the other hand, as rotating the image will also rotate the derivative distribution without changing the shape, Eq. (\ref{Eq:dpq}) is rotation-invariant.} The drawback is also obvious: Eq. (\ref{Eq:dpq}) cannot distinguish some shapes like corners from edges as the entire spreading condition of the distribution is evaluated only. However, since the proposed detector (Section \ref{sec:EAS}) does not depend on specific geometric shapes like corners or edges, it will not be a practical issue in this work.


\begin{figure}[t]
	\centering
\subfloat[]{\raisebox{2ex}
{\includegraphics[width=0.3\linewidth]{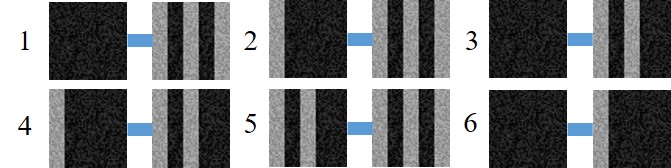}}
}
\subfloat[]{
\includegraphics[width=0.3\linewidth]{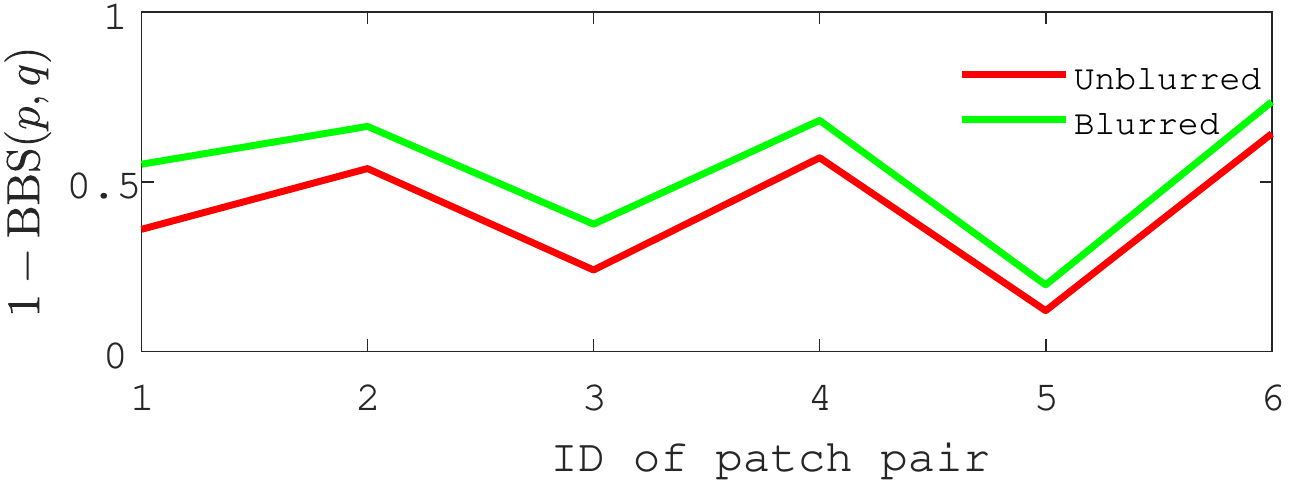}
}
\subfloat[]{
\includegraphics[width=0.3\linewidth]{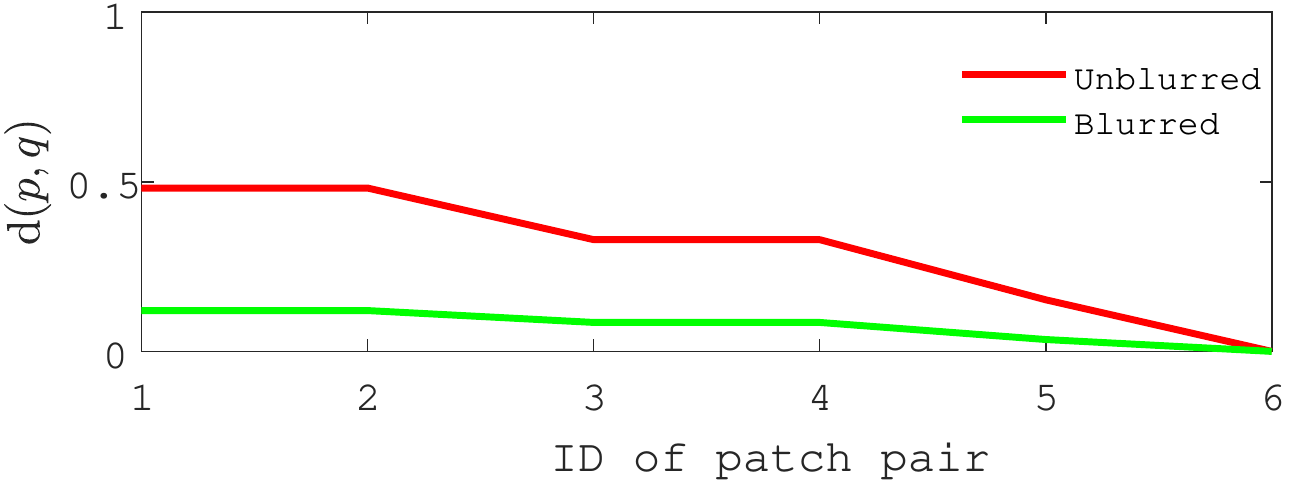}
}
 \caption{(a) Patch pairs are numbered in the descending order in terms of the difference of the number of edges in two patches from 1 to 6 (i.e., a smaller difference indicates a greater number). (b) Distance between two patches' derivative distributions calculated by BBS \cite{dekel2015best}. (c) Distance between two patches' derivative distributions calculated with $\mathrm{d}(p,q)$.  The scores (i.e., distances computed by Eq. (\ref{Eq:dpq})) are required to be non-ascending with the increase of pair number: $\mathrm{d}(p_1,q_1)\geq \mathrm{d}(p_2,q_2)\geq...\geq\mathrm{d}(p_6,q_6)$.}
\label{fig:distance}
\end{figure}

Fig. \ref{fig:patchtest} illustrates a simple example which displays four noisy patches (first row) and their blurred ones (second row) with different appearances. $\lambda_{min}$ and $\lambda_{max}$ of each patch's covariance matrix are calculated and shown at the same time. As the result of blurring, eigenvalues decrease. Note that as the patches in the fourth column are generated by rotating the patches in the third column 90 degrees clockwise, and the distributions are also rotated, with eigenvalues unchanged. The distance measurement is supposed to be able to quantify the visual difference with respect to the edges brought by the white strips, and invariant to rotations. Fig. \ref{fig:distance}(b) shows the change of distance scored by BBS \cite{dekel2015best} with the decrease of visual similarity between patches shown in Fig. \ref{fig:distance}(a). In this example, BBS fails to evaluate the visual similarity properly with rotated patches. On the other hand, in Fig.\ref{fig:distance}(c), despite the entire scale of distances decreases due to motion blur, the magnitude relationship between any two patch sets is preserved. $d(p,q)$ is capable to measure the visual difference properly under both blurred and unblurred situations.

With regard to keypoint detection, calculating the eigenvalues of covariance matrices over numerous image patches can be highly time-consuming. To mitigate this issue, the covariance matrix $\Sigma$ can be simplified by further assuming the mean $I_x$ and mean $I_y$ of $p$ and $q$ are zeros:
\begin{equation}
\Sigma=
  \begin{bmatrix}
    \mathrm{E}[(I_x-\mu_x)(I_x-\mu_x)] & \mathrm{E}[(I_x-\mu_x)(I_y-\mu_y)]\\
    \mathrm{E}[(I_x-\mu_x)(I_y-\mu_y)] & \mathrm{E}[(I_y-\mu_y)(I_y-\mu_y)]
  \end{bmatrix}\sim
  \begin{bmatrix}
    \mathrm{E}(I_x^2) & \mathrm{E}(I_xI_y)\\
    \mathrm{E}(I_xI_y) & \mathrm{E}(I_y^2)
  \end{bmatrix}.
\end{equation}
As 
\begin{equation}
\lambda_{min}+\lambda_{max}=\mathrm{trace}(\Sigma),
\end{equation}
then $\mathrm{d}(p,q)$ can be further rewritten as
\begin{equation}
\mathrm{d}(p,q) = \vert {I_x^p}^2-{I_x^q}^2+{I_y^p}^2-{I_y^q}^2 \vert,
\label{Eq:dpqsimple}
\end{equation}
where ${I_x^p}^2$ represents the expectation of $I_x^2$ with respect to patch $p$, and other variables are in a similar manner. Note that in Eq. (\ref{Eq:dpqsimple}) only derivatives need to be calculated to estimate the distance between two patches' derivative distributions, thus it can be efficient. Greater values of $\mathrm{d}(p,q)$ contribute more to the asymmetry. Based on the above observation and analysis, we will explain how to use $\mathrm{d}(p,q)$ to model the asymmetry in Section \ref{sec:EAS}.

\subsection{Eigenvalue Asymmetry (EAS)}
\label{sec:EAS}
The degradation caused by image blur will decrease the intensity variations of the original image, thus introducing more uniform regions which are naturally with high degree of symmetry. Taking this factor into consideration, we measure asymmetry instead of symmetry. We claim that the regions with larger $\mathrm{d}(p,q)$ in different directions maintain better distinctiveness than other regions.\ignore{In other words, the pixel with high EAS score will accidentally coincide with other pixels less frequently.} A radius parameter $r$ is introduced to test the asymmetry at pixel $I(i,j)$ radially. Specifically,
\begin{equation}
(P^{i,j,r},Q^{i,j,r})=\mathrm{R}(i,j,r),
\end{equation}
where $\textrm{R}$ defines a function that generates two patch sets in the same size $N$ with respect to $r$. $P^{i,j,r}_n \in P^{i,j,r}$ and $Q^{i,j,r}_n \in Q^{i,j,r}$ ($n \in [ 1,N ]$) are spatially symmetric across the coordinates $(i,j)$, representing image patches from two sets, respectively. By averaging the distance defined in Eq. (\ref{Eq:dpqsimple}) over pairs of patches $(P^{i,j,r}_n,Q^{i,j,r}_n)$, the metric for evaluating the asymmetry can be defined as follows,
\begin{equation}
\mathrm{EAS}(i,j,r) = \frac{1}{N}\sum_{n=1}^{N}\mathrm{d}(P^{i,j,r}_n,Q^{i,j,r}_n).
\label{Eq:EAS}
\end{equation}
Theoretically, $N$ equals to $\pi r$ and a higher EAS score means larger asymmetry. By thresholding EAS, we can detect top scored pixels as keypoints. Since we estimate asymmetry radially, the EAS also holds the rotation-invariance.

\subsection{Scale Space}
\begin{figure}[t]
	\centering
\subfloat[Guassian blur ($\sigma=1$)]{
{\includegraphics[width=0.95\linewidth]{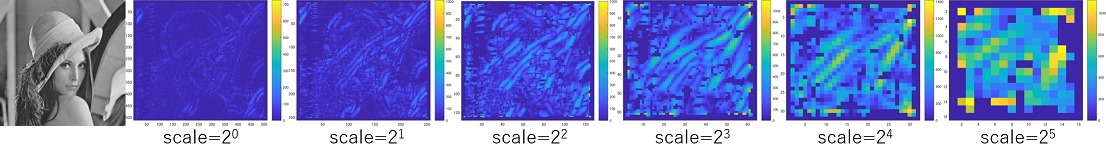}}
 }
 
\subfloat[Guassian blur ($\sigma=9$)]{
\includegraphics[width=0.95\linewidth]{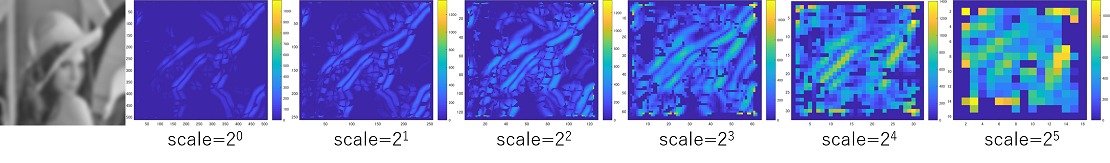}
}
\caption{Score maps of EAS for each octave under different Gaussian blur. }
\label{fig:scale1scale2}
\end{figure}

To deal with heavy blur as well as image scaling, the proposed detector must be able to evaluate each pixel in scale space. We use the kernel described in \cite{burt1987laplacian} to generate octaves for building the Gaussian image pyramid by a scaling factor of 0.5 (i.e., the size of image is halved from previous octave to the next octave). We use the concept of blob to describe a keypoint, with a changeable radius $r$ (equivalent to the radius in Eq. (\ref{Eq:EAS})) representing its scale. As suggested by many previous works, dividing octaves into layers and applying Laplacian-of-Gaussian (LoG) or Gaussian-of-Difference (DoG) is a conventional scheme for scale search of keypoints. We do not use it to find the local maximum of EAS in scale space, because of three reasons: (1) besides the additional calculation of EAS introduced by layers in each octave, either convolving LoG kernel or calculating DoG induces significant computational overhead, especially for images of large size. (2) $r$ for highest EAS response is actually hard to be determined. Because the values of EAS with different $r$ are not guaranteed to be a unimodal distribution, which is unlike filter based methods. (3) In spite of the possible requirements for precise $r$ during the procedure of feature descriptor extraction, we found in experiments that the discrete values ($r$ = $2^0$, $2^1$, $2^2$, $...$) are sufficient to make keypoints distinctive. In Fig. \ref{fig:scale1scale2}, we show each octave's score map of EAS. Basically, EAS tends to give higher score to pixels nearby edges as higher gradient variation can usually be observed. It is worth pointing out that with the increase of scale level, though encountering very different degrees of blur, the maps become similar with each other since the third octave. This demonstrates our EAS is robust to blur.

\begin{figure}[t]
	\centering
\subfloat[]{\raisebox{9ex}
{\includegraphics[width=0.2\linewidth]{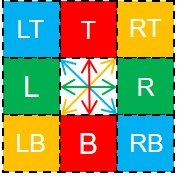}}
 }
\subfloat[]{
\includegraphics[width=0.55\linewidth]{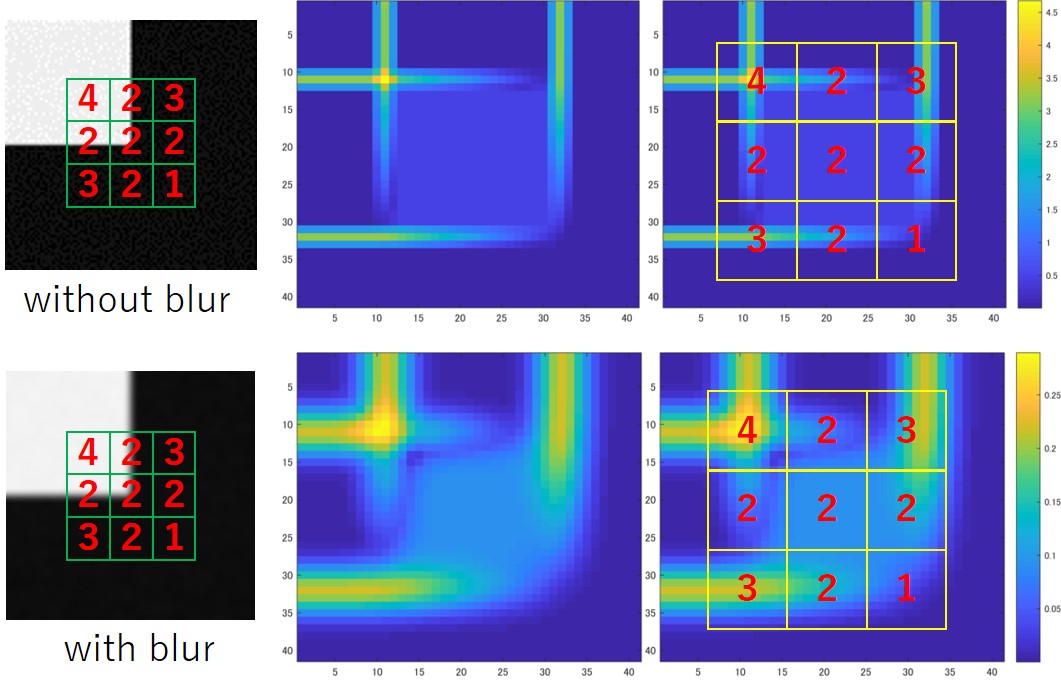}
}
\subfloat[]{\raisebox{1ex}
{\includegraphics[width=0.215\linewidth]{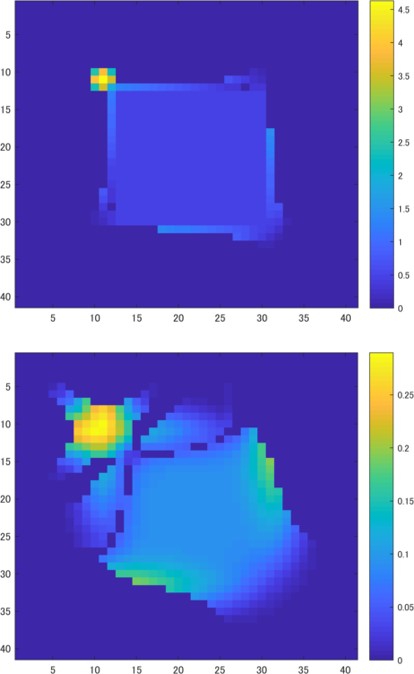}
}}
\caption{\textbf{(a)} EAS is calculate by averaging $\mathrm{d}(\mathrm{LT},\mathrm{RB})$, $\mathrm{d}(\mathrm{L},\mathrm{R})$, $\mathrm{d}(\mathrm{LB},\mathrm{RT})$ and $\mathrm{d}(\mathrm{T},\mathrm{B})$, where eight neighbors are used. \textbf{(b)} EAS around a corner. On the right side of (b), we use the number of directions with distinct derivative distributions (i.e., 1-4) to show the EAS differences in the $3\times3$ grids, which is consistent with human visual observations (left side of (b)). A larger number denotes a larger EAS. \textbf{(c)} Results after edge effect elimination.}
\label{fig:cornerESA}
\end{figure}

\subsection{Implementation Details}
In this section, we introduce the implementation details in order to improve the reproducibility. 

\textbf{EAS with partial distributions.} Heavy noises would enlarge the shape of derivative distribution. To alleviate this issue, in the implementation we only use partial distributions (e.g., points in the first quadrant) to calculate the EAS and set a upper bound (e.g., 1.0) for squared derivatives.

\textbf{Selection of $r$.} We simply use the radial asymmetry test shown in Fig. \ref{fig:cornerESA}(a) with eight neighbors (i.e., approximately equivalent to $r=1$ in the first octave). Note that the $r$ is related to the octave number. For example, in the second octave, as we use Fig. \ref{fig:cornerESA}(a) to test with the halved image, $r$ is $2^1$. A smaller $r$ is more robust to space-variant blur as the EAS is computed more locally. Fig. \ref{fig:cornerESA}(b) shows an example of EAS calculation around a corner. In the case of EAS, before visualizing the EAS map, we can observe and predict that the pixels inside the corner (left-top direction) would have the highest responses while pixels outside (right-bottom direction) would have the lowest responses. The numbers from 1 to 4 indicate the number of directions along which different derivatives can be visually observed with respect to each position. The actual EAS map shown on right side well confirms the above observation with both unblurred/blurred corners. 

\textbf{Non-maximum suppression.} After calculating EAS in each octave, we represent each keypoint with a blob of radius $r$ centered at $(i, j)$ in the original image. We then select a subset of keypoints that are locally strong at each octave (e.g., within 3$\times$3 neighbors), allowing keypoints with the same $(i, j)$ but different $r$ possibly appear at the same time. 

\textbf{Elimination of edge effect.} The points near edges potentially have steep gradient changes and thus yield in higher EAS scores, and have the property: $\lambda_{max}\gg\lambda_{min}$. To eliminate such points and achieve more distinctive keypoints, we exclude the pixels when $\lambda_{max}/\lambda_{min}>thr$ (e.g., $thr=5$). Fig. \ref{fig:cornerESA}(c) shows an example of edge effect elimination. Comparing to (b), the EAS responses due to the asymmetry of edges are eliminated.

\textbf{Average of the squared derivatives.} Expectation of squared derivatives, mentioned in Eq. \ref{Eq:dpqsimple}, can be only calculated from the center pixel. To improve the robustness, we average the squared derivatives in a $K\times K$ neighborhood, with $K$ in each octave determined by the scale of keypoints. In our implementation, we model $K=10/r$ and set $r=2^0,..., 2^5$.

\section{Experimental Results}
In this section, we compare our proposed method with state of the art detectors quantitatively and qualitatively. We validate our method in a variety of scenarios, including space-invariant blur, space-variant blur, complex blur, affine transformation, scaling and noise. To show the robustness of our approach, all the involved parameters are fixed for all experiments.
\subsection{Experimental Setting}
It is difficult to achieve ground truth for blurred real data.\ignore{there exists no ideal blur-invariant feature descriptor to the best of our knowledge.} To make fair and accurate comparisons focusing on keypoint detection, we use the images from the widely known dataset \cite{mikolajczyk2005performance} and an additional $512\times512$ ``Lena'' image to generate synthetic test data.

\begin{figure}[t]
	\centering
\subfloat{
\includegraphics[width=0.23\linewidth]{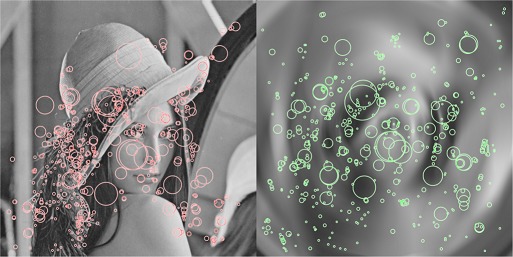}
 }
\subfloat{
\includegraphics[width=0.23\linewidth]{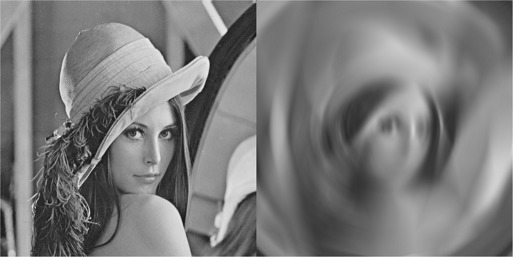}
}
\subfloat{
\includegraphics[width=0.23\linewidth]{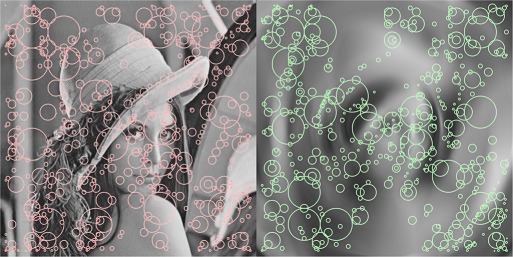}
}
\subfloat{
\includegraphics[width=0.23\linewidth]{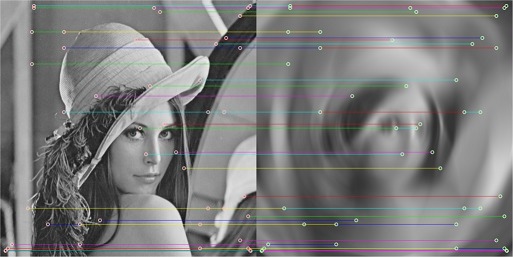}
}
\setcounter{subfigure}{0}
\subfloat[Fast-Hessian]{
\includegraphics[width=0.23\linewidth]{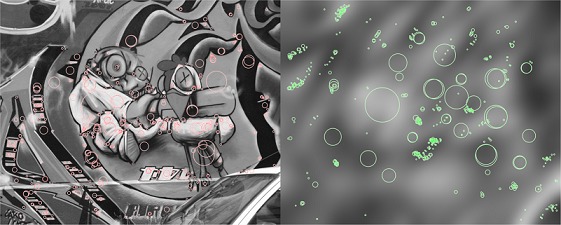}
 }
\subfloat[Fast-Hessian]{
\includegraphics[width=0.23\linewidth]{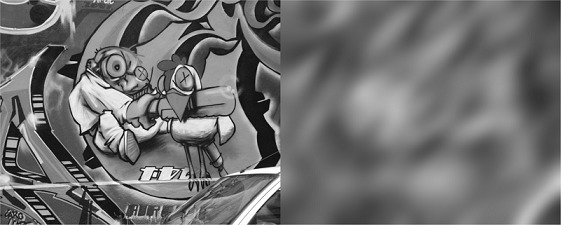}
}
\subfloat[Our method]{
\includegraphics[width=0.23\linewidth]{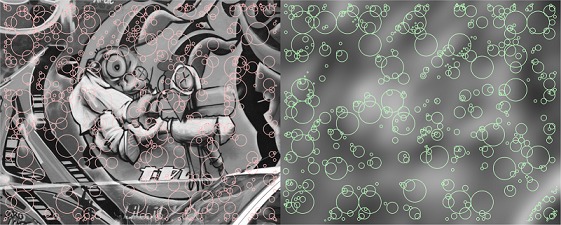}
}
\subfloat[Our method]{
\includegraphics[width=0.23\linewidth]{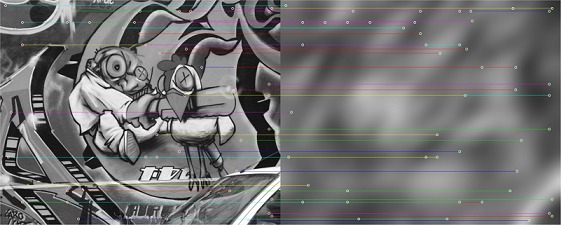}
}

\caption{Comparative results with complex blur. ``Lena'' (first row): the complex blur is generated in the sequence of applying rotational blur ($\pi/6$ degree), motion blur ($l=30,\theta=\pi/2$) and Gaussian blur ($\sigma=9$). ``Graffiti'' (second row): the complex blur is generated in the sequence of applying strong motion blur ($l=100,\theta=\pi/4$) and Gaussian blur ($\sigma=20$).}
\label{fig:combine}
\end{figure}

\begin{figure}[t]
	\centering
\subfloat{
\includegraphics[width=0.23\linewidth]{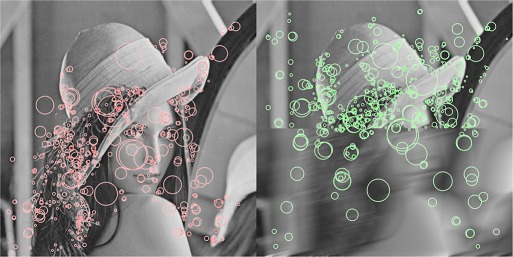}
 }
\subfloat{
\includegraphics[width=0.23\linewidth]{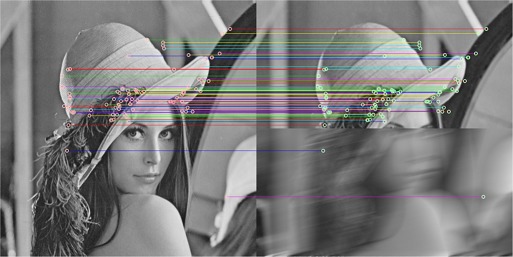}
}
\subfloat{
\includegraphics[width=0.23\linewidth]{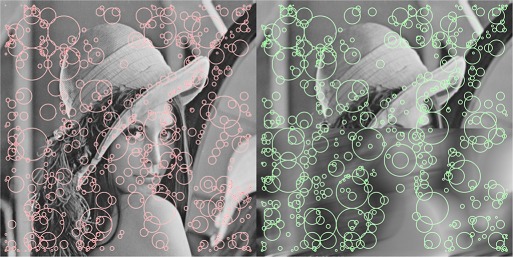}
}
\subfloat{
\includegraphics[width=0.23\linewidth]{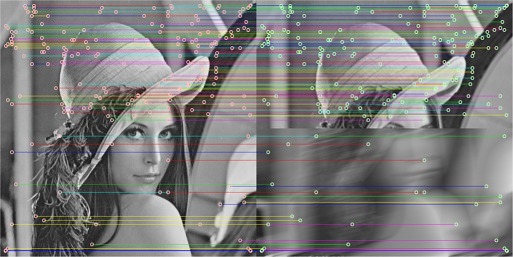}
}
\setcounter{subfigure}{0}
\subfloat[Fast-Hessian]{
\includegraphics[width=0.23\linewidth]{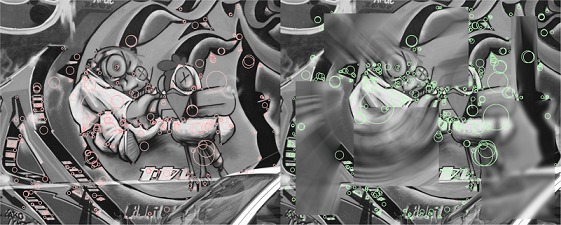}
 }
\subfloat[Fast-Hessian]{
\includegraphics[width=0.23\linewidth]{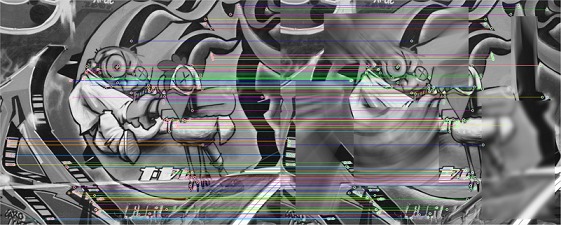}
}
\subfloat[Our method]{
\includegraphics[width=0.23\linewidth]{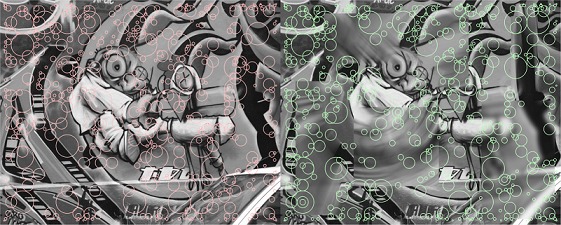}
}
\subfloat[Our method]{
\includegraphics[width=0.23\linewidth]{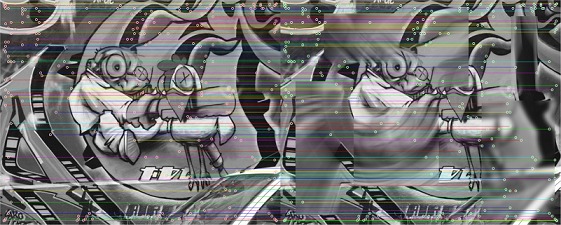}
}

\caption{Comparative results with space-variant blur. ``Lena'' (first row): motion blur is applied in the lower half. ``Graffiti'' (second row): random rectangular regions are selected to apply random types of blur, including Gaussian, radial, motion and rotational blur. }
\label{fig:variant}
\end{figure}

\begin{figure}[t]
	\centering
\subfloat{
\includegraphics[width=0.23\linewidth]{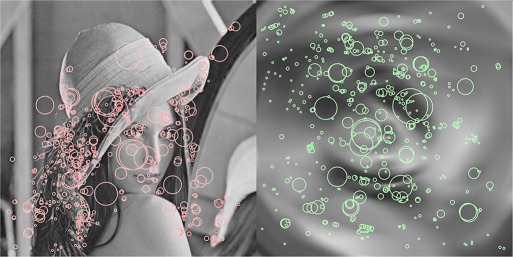}
 }
\subfloat{
\includegraphics[width=0.23\linewidth]{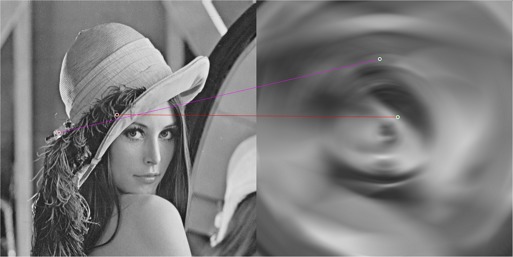}
}
\subfloat{
\includegraphics[width=0.23\linewidth]{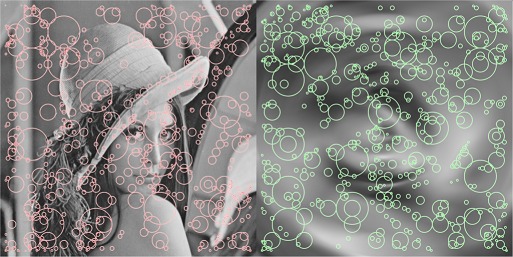}
}
\subfloat{
\includegraphics[width=0.23\linewidth]{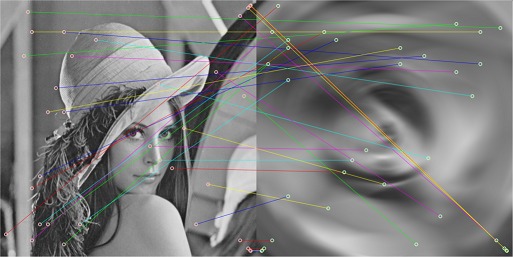}
}
\setcounter{subfigure}{0}
\subfloat[Fast-Hessian]{
\includegraphics[width=0.23\linewidth]{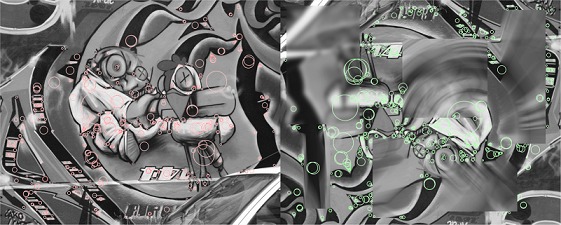}
 }
\subfloat[Fast-Hessian]{
\includegraphics[width=0.23\linewidth]{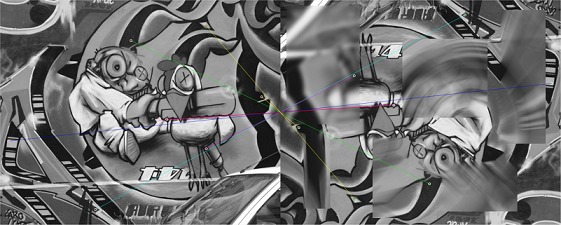}
}
\subfloat[Our method]{
\includegraphics[width=0.23\linewidth]{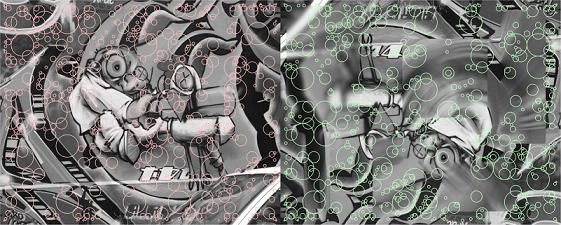}
}
\subfloat[Our method]{
\includegraphics[width=0.23\linewidth]{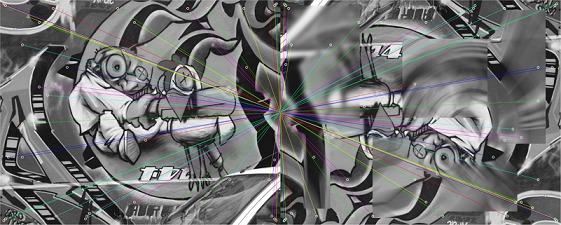}
}

\caption{Comparative results with rotation. The ``Lena'' example from Fig. \ref{fig:combine} and ``Graffiti'' example from Fig. \ref{fig:variant} are rotated by $\pi/2$ and $\pi$ clockwise, respectively. }
\label{fig:rotation}
\end{figure}

\begin{figure}[t]
	\centering
\subfloat{
\includegraphics[width=0.23\linewidth]{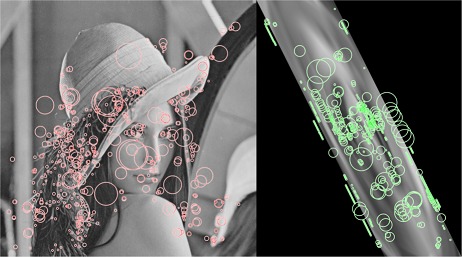}
 }
\subfloat{
\includegraphics[width=0.23\linewidth]{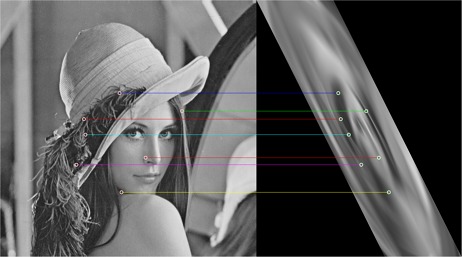}
}
\subfloat{
\includegraphics[width=0.23\linewidth]{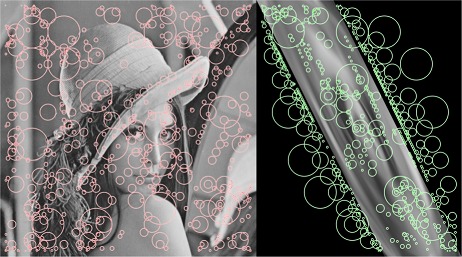}
}
\subfloat{
\includegraphics[width=0.23\linewidth]{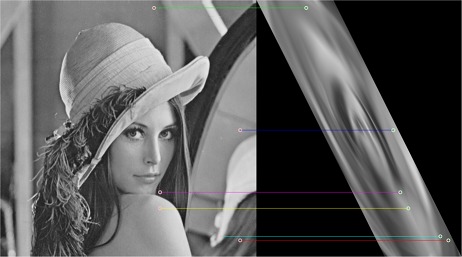}
}
\setcounter{subfigure}{0}
\subfloat[Fast-Hessian]{
\includegraphics[width=0.23\linewidth]{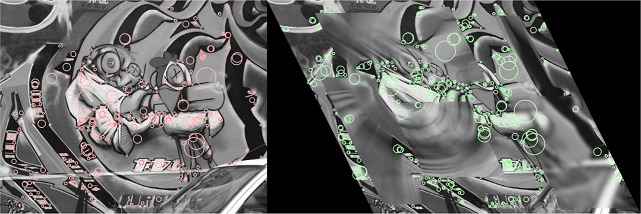}
 }
\subfloat[Fast-Hessian]{
\includegraphics[width=0.23\linewidth]{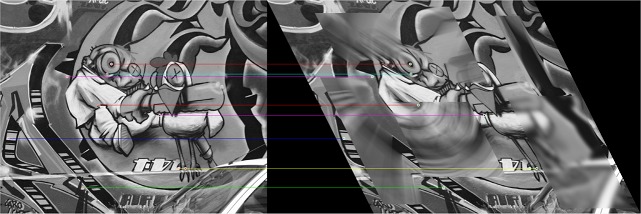}
}
\subfloat[Our method]{
\includegraphics[width=0.23\linewidth]{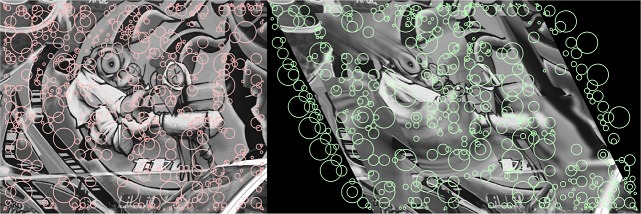}
}
\subfloat[Our method]{
\includegraphics[width=0.23\linewidth]{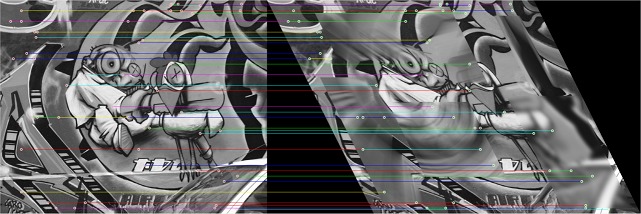}
}

\caption{Comparative results with affine transformation. The ``Lena'' example from Fig. \ref{fig:combine} and ``Graffiti'' example from Fig. \ref{fig:variant} are transformed by the matrices $[1\ 0\ 0; 0.5\ 1\ 0; 0\ 0\ 1]$ and $[0.3\ 0\ 0; 0.5\ 1\ 0; 0.5\ 0\ 1]$, respectively.}
\label{fig:affine}
\end{figure}

\begin{figure}[t]
	\centering
  \subfloat{
\includegraphics[width=0.23\linewidth]{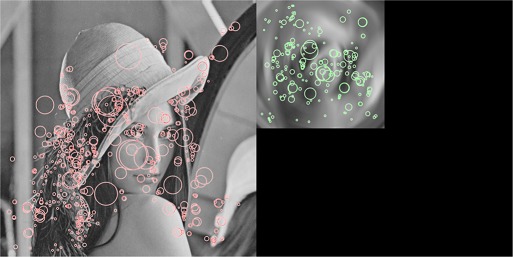}
}
\subfloat{
\includegraphics[width=0.23\linewidth]{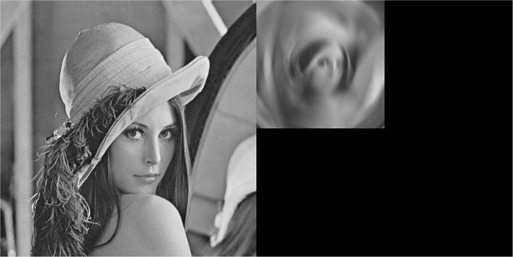}
}
\subfloat{
\includegraphics[width=0.23\linewidth]{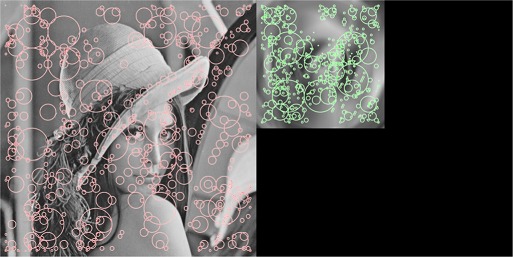}
 }
\subfloat{
\includegraphics[width=0.23\linewidth]{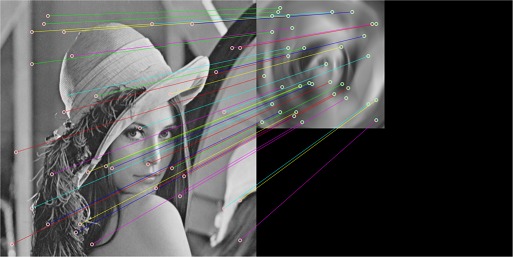}
}
\setcounter{subfigure}{0}
\subfloat[Fast-Hessian]{
\includegraphics[width=0.23\linewidth]{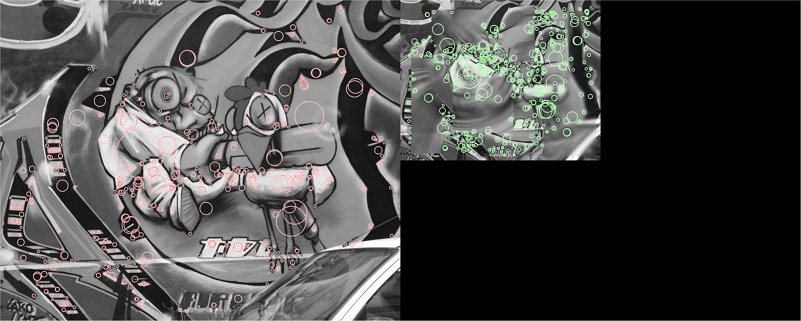}
 }
\subfloat[Fast-Hessian]{
\includegraphics[width=0.23\linewidth]{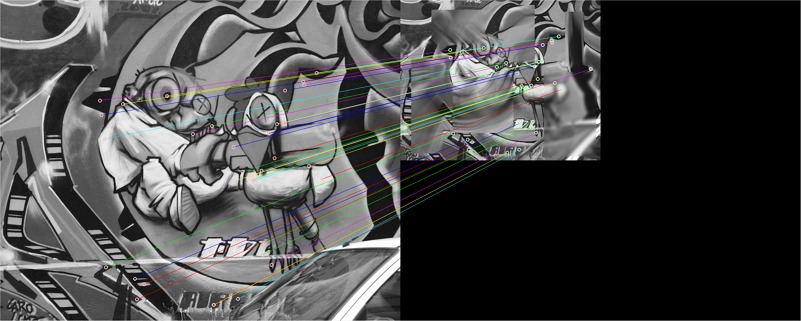}
}
\subfloat[Our method]{
\includegraphics[width=0.23\linewidth]{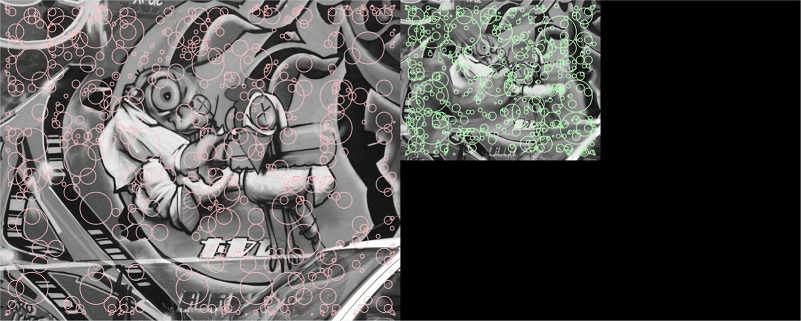}
}
\subfloat[Our method]{
\includegraphics[width=0.23\linewidth]{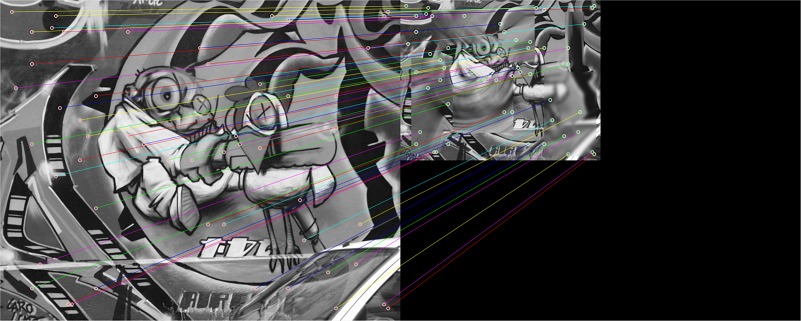}
}
\caption{Comparative results with scaling. The ``Lena'' example from Fig. \ref{fig:combine} and ``Graffiti'' example from Fig. \ref{fig:variant} are both zoomed out by a factor of $0.5$.}
\label{fig:scale}
\end{figure}

\begin{figure}[t]
	\centering
  \subfloat{
\includegraphics[width=0.23\linewidth]{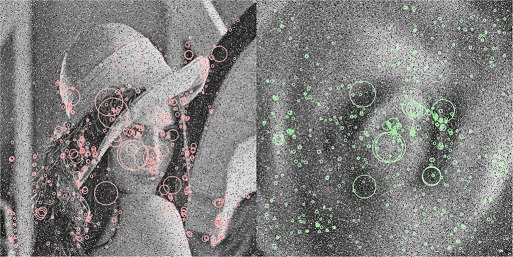}
}
\subfloat{
\includegraphics[width=0.23\linewidth]{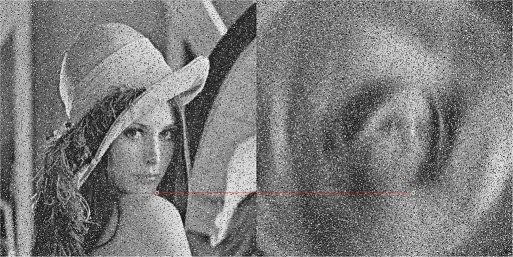}
}
\subfloat{
\includegraphics[width=0.23\linewidth]{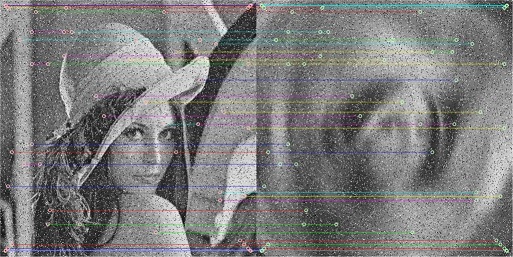}
 }
\subfloat{
\includegraphics[width=0.23\linewidth]{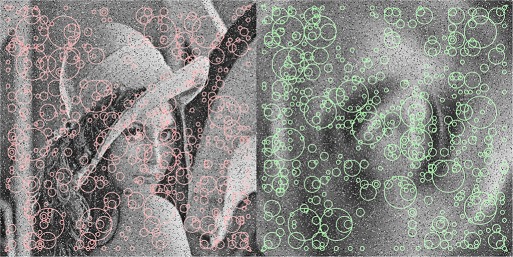}
}

\setcounter{subfigure}{0}
\subfloat[Fast-Hessian]{
\includegraphics[width=0.23\linewidth]{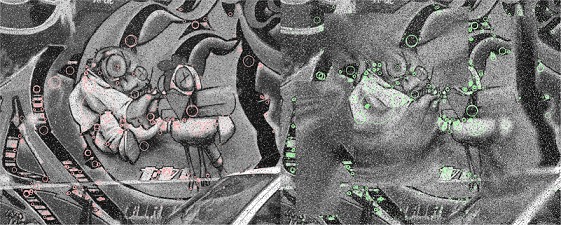}
 }
\subfloat[Fast-Hessian]{
\includegraphics[width=0.23\linewidth]{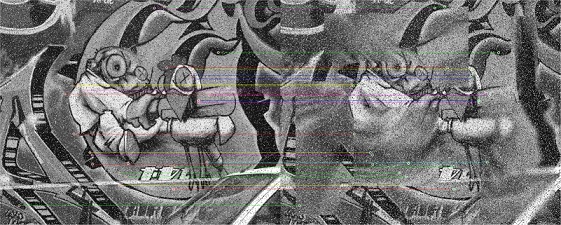}
}
\subfloat[Our method]{
\includegraphics[width=0.23\linewidth]{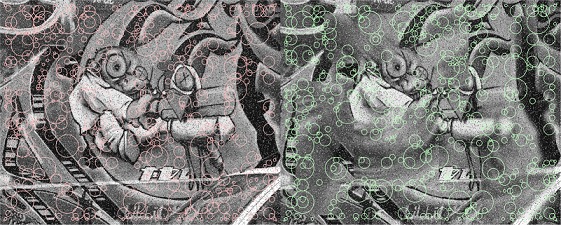}
}
\subfloat[Our method]{
\includegraphics[width=0.23\linewidth]{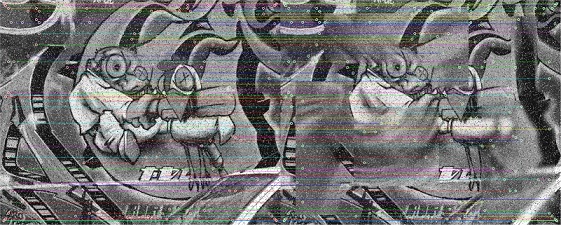}
}

\caption{Comparative results with salt and pepper noise. 10\% random pixels of the ``Lena'' example from Fig. \ref{fig:combine} and ``Graffiti'' example from Fig. \ref{fig:variant} are replaced by salt and pepper noise (with different random seeding). }
\label{fig:noise}
\end{figure}

As suggested by previous works \cite{mikolajczyk2002affine,schmid2000evaluation}, repeatability rate is used for evaluation. Only if two keypoints from two images are located at the same relative position by considering the geometric transformation, we view these two points as ``corresponding points''. $N_c$ is used to represent the number of corresponding points. The performance of a detector is usually sensitive to threshold tuning, which may bring unfairness to comparisons. In our experiments, $TopN$ keypoints (with high confidence) in each image are thus selected for evaluation after ranking all the candidate keypoints. We redefine the repeatability as $N_c/TopN$. From Fig. \ref{fig:combine} to \ref{fig:noise}, the $TopN$ is fixed to 500.

State of the art detectors are selected for comparisons, including Fast-Hessian from SURF \cite{bay2006surf}, DoG from SIFT \cite{lowe2004distinctive}, Harris corner \cite{harris1988combined}, FAST \cite{rosten2010faster}, minimum eigenvalue \cite{shi1994good} and BRISK \cite{leutenegger2011brisk} with the same number of octaves. The metric score threshold is relaxed to generate sufficient candidates, and $TopN$ of them are selected for evaluation. Affine region detectors are not compared as we aim to analyze the performance under image blur rather than geometric transformations. In Section 4.2, Fast-Hessian is chosen for comparison because its superiority stated in \cite{bay2006surf}.



\subsection{Qualitative Results}
Fig. \ref{fig:combine} to Fig. \ref{fig:noise} show the qualitative comparative results under complex blur, space-invariant blur, blur+rotation, blur+Affine transformation, blur+scaling, blur+noise respectively. Each red circle represents a keypoint with its supporting radius $r$. For clearness, corresponding points are connected by lines in different colors. The repeatability of all figures is listed in Table \ref{table1}. As observed both visually and quantitatively, our method outperforms the state of the art detector in most scenarios. However, we notice two failures: (1) In the case of ``Lena'' in Fig. \ref{fig:affine}, a smaller $N_c$ is detected by EAS. One possible reason is that the padded image produces additional boundaries on both sides that have strong EAS responses. Besides, EAS may be inappropriate for the sheering operation which compresses the pixels and changes the spatial relationships between pixels. (2) In Fig. \ref{fig:variant}, although EAS detects more keypoints in blurred regions comparing with Fast-Hessian, the amounts are unbalanced between blurred and unblurred regions. We suspect that the EAS scores calculated from unblurred regions are usually higher because of steeper gradients. 



\begin{table}[h]%
 \centering
\caption{Repeatability of Fig. \ref{fig:combine} to \ref{fig:noise}.}

 \begin{tabular}{|l|l|l|} \hline
  \begin{tabular}{c}\ \ Method\end{tabular} & \begin{tabular}{c}\ \ Fig.\end{tabular} & \begin{tabular}{c}Repeatability \\ (``Lena'', ``Graffiti'')\end{tabular}\\ \hline \hline
   & Fig. \ref{fig:combine} & 0/500, 0/500 \\ \cline{2-3}
   & Fig. \ref{fig:variant} & 88/500, 170/500 \\ \cline{2-3}
Fast-Hessian & Fig. \ref{fig:rotation} & 2/500, 9/500 \\ \cline{2-3}
   & Fig. \ref{fig:affine} & \textbf{7/500}, 9/500 \\ \cline{2-3}
   & Fig. \ref{fig:scale} & 0/500, 42/500 \\ \cline{2-3}
   & Fig. \ref{fig:noise} & 1/500, 36/500 \\ \hline
   
   & Fig. \ref{fig:combine} & \textbf{48/500, 50/500} \\ \cline{2-3}
   & Fig. \ref{fig:variant} & \textbf{195/500, 263/500} \\ \cline{2-3}
Our method & Fig. \ref{fig:rotation} & \textbf{35/500, 52/500} \\ \cline{2-3}
   & Fig. \ref{fig:affine} & 6/500, \textbf{54/500} \\ \cline{2-3}
   & Fig. \ref{fig:scale} & \textbf{37/500, 58/500} \\ \cline{2-3}
   & Fig. \ref{fig:noise} & \textbf{57/500, 180/500} \\ \hline
 \end{tabular}
 \label{table1}%
\end{table}

\begin{figure}[h]
	\centering
\subfloat[]{
\includegraphics[width=0.5\linewidth]{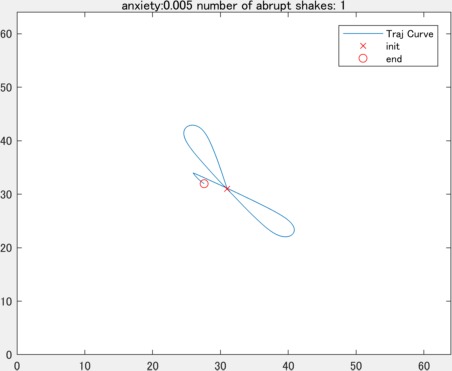}
 }
\subfloat[]{
\includegraphics[width=0.1\linewidth]{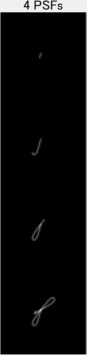}
}

\subfloat[SURF]{
\includegraphics[width=0.23\linewidth]{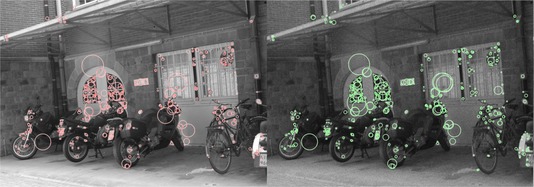}
}
\subfloat[$N_c=8$ (SURF)]{
\includegraphics[width=0.23\linewidth]{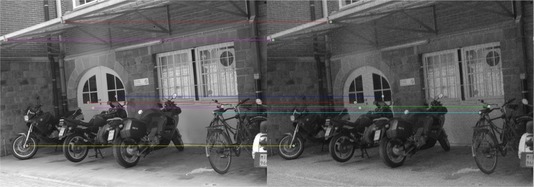}
}
\subfloat[EAS]{
\includegraphics[width=0.23\linewidth]{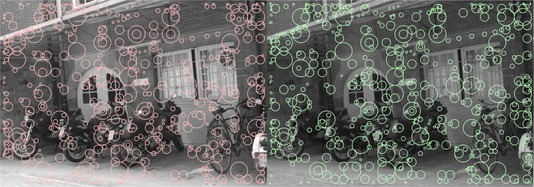}
}
\subfloat[$N_c=254$ (EAS)]{
\includegraphics[width=0.23\linewidth]{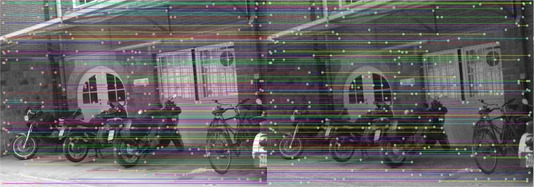}
}

\subfloat[SURF]{
\includegraphics[width=0.23\linewidth]{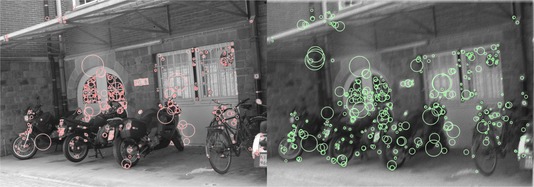}
}
\subfloat[$N_c=0$ (SURF)]{
\includegraphics[width=0.23\linewidth]{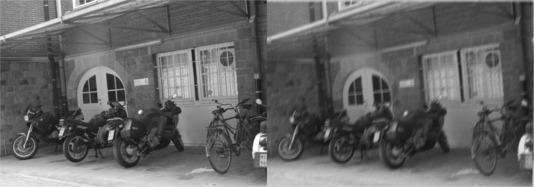}
}
\subfloat[EAS]{
\includegraphics[width=0.23\linewidth]{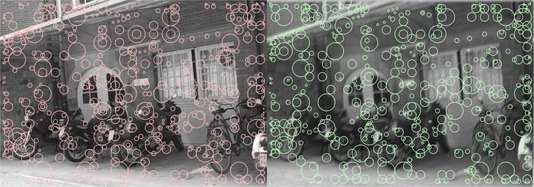}
}
\subfloat[$N_c=153$ (EAS)]{
\includegraphics[width=0.23\linewidth]{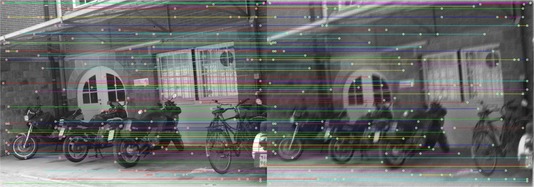}
}

\subfloat[SURF]{
\includegraphics[width=0.23\linewidth]{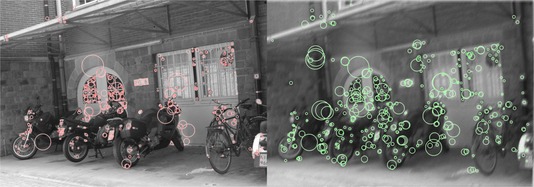}
}
\subfloat[$N_c=0$ (SURF)]{
\includegraphics[width=0.23\linewidth]{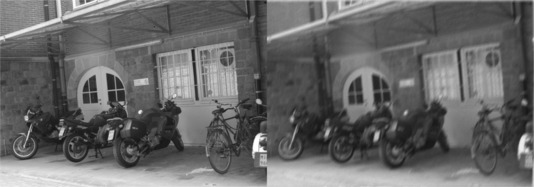}
}
\subfloat[EAS]{
\includegraphics[width=0.23\linewidth]{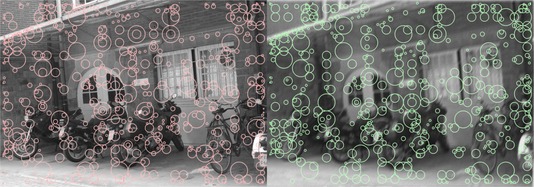}
}
\subfloat[$N_c=182$ (EAS)]{
\includegraphics[width=0.23\linewidth]{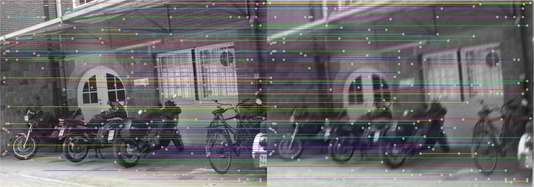}
}

\subfloat[SURF]{
\includegraphics[width=0.23\linewidth]{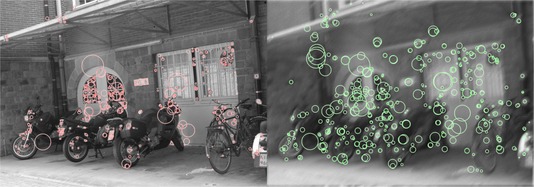}
}
\subfloat[$N_c=3$ (SURF)]{
\includegraphics[width=0.23\linewidth]{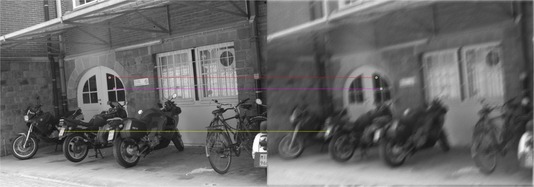}
}
\subfloat[EAS]{
\includegraphics[width=0.23\linewidth]{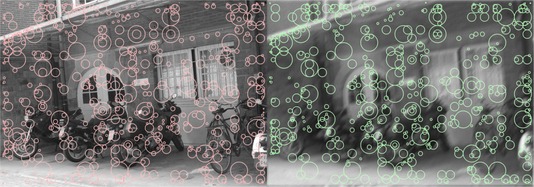}
}
\subfloat[$N_c=218$ (EAS)]{
\includegraphics[width=0.23\linewidth]{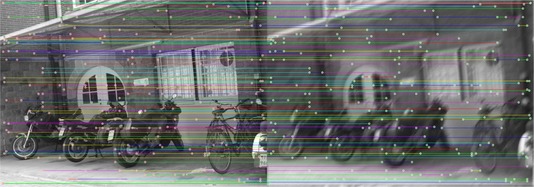}
}
\caption{Keypoint detection results on a Bike image.}
\label{supp1}
\end{figure}

\begin{figure}[h]
	\centering
\subfloat[]{
\includegraphics[width=0.5\linewidth]{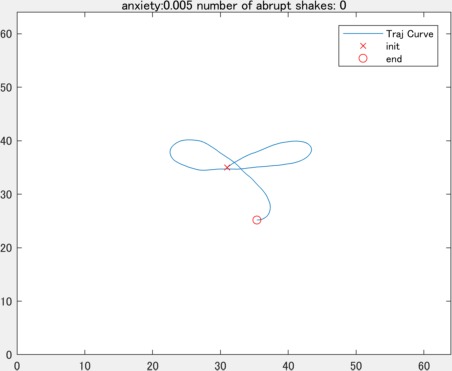}
 }
\subfloat[]{
\includegraphics[width=0.1\linewidth]{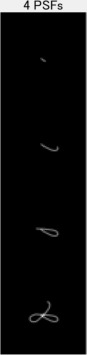}
}

\subfloat[SURF]{
\includegraphics[width=0.23\linewidth]{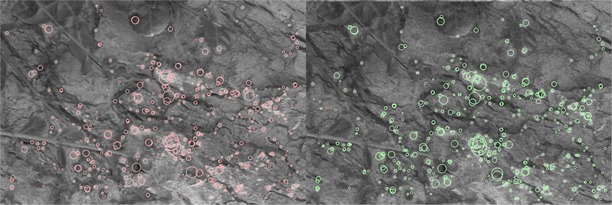}
}
\subfloat[$N_c=1$ (SURF)]{
\includegraphics[width=0.23\linewidth]{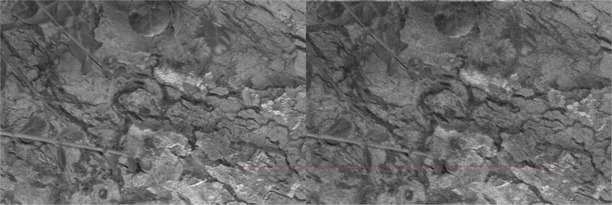}
}
\subfloat[EAS]{
\includegraphics[width=0.23\linewidth]{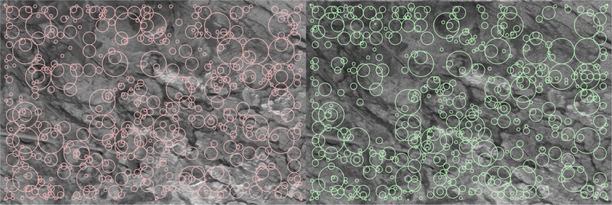}
}
\subfloat[$N_c=214$ (EAS)]{
\includegraphics[width=0.23\linewidth]{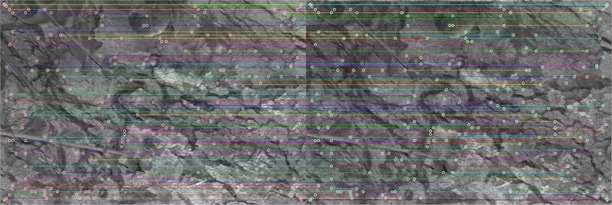}
}

\subfloat[SURF]{
\includegraphics[width=0.23\linewidth]{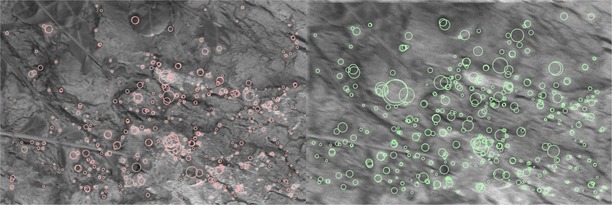}
}
\subfloat[$N_c=1$ (SURF)]{
\includegraphics[width=0.23\linewidth]{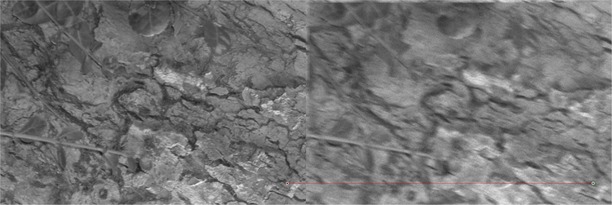}
}
\subfloat[EAS]{
\includegraphics[width=0.23\linewidth]{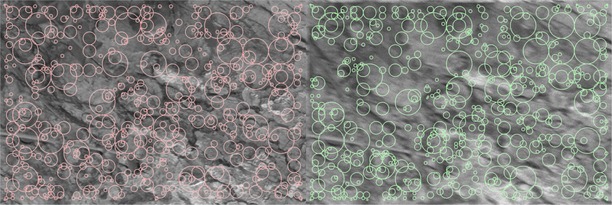}
}
\subfloat[$N_c=118$ (EAS)]{
\includegraphics[width=0.23\linewidth]{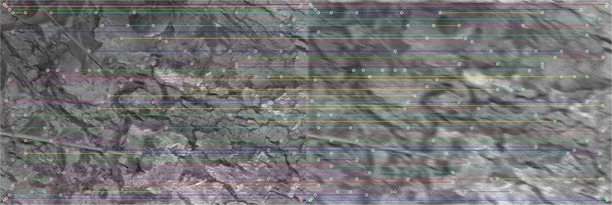}
}

\subfloat[SURF]{
\includegraphics[width=0.23\linewidth]{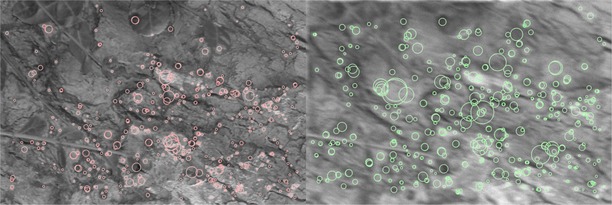}
}
\subfloat[$N_c=1$ (SURF)]{
\includegraphics[width=0.23\linewidth]{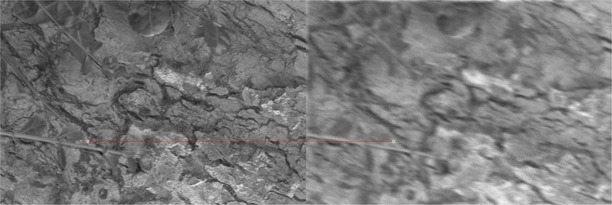}
}
\subfloat[EAS]{
\includegraphics[width=0.23\linewidth]{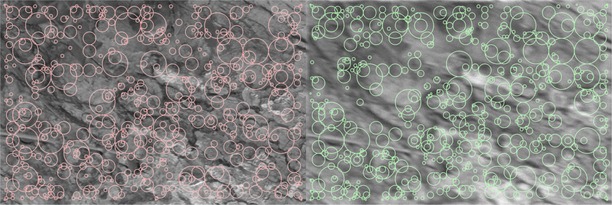}
}
\subfloat[$N_c=144$ (EAS)]{
\includegraphics[width=0.23\linewidth]{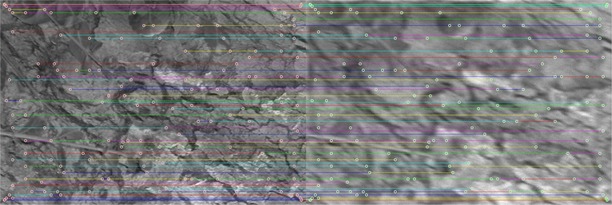}
}

\subfloat[SURF]{
\includegraphics[width=0.23\linewidth]{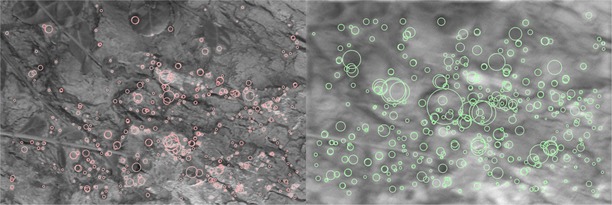}
}
\subfloat[$N_c=2$ (SURF)]{
\includegraphics[width=0.23\linewidth]{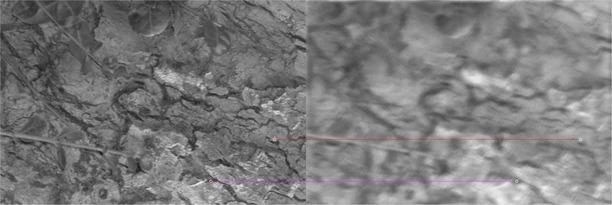}
}
\subfloat[EAS]{
\includegraphics[width=0.23\linewidth]{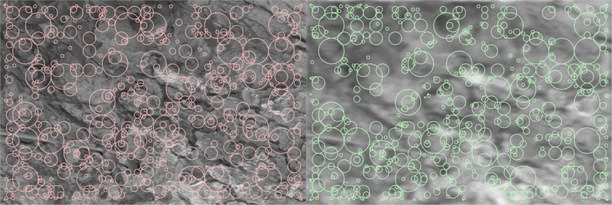}
}
\subfloat[$N_c=205$ (EAS)]{
\includegraphics[width=0.23\linewidth]{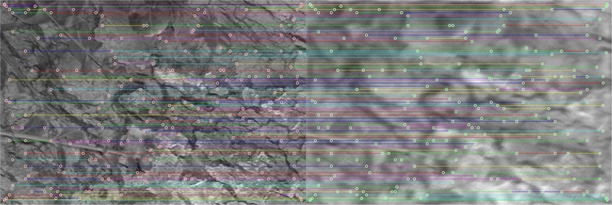}
}
\caption{Keypoint detection results on a Bark image.}
\label{supp2}
\end{figure}

\begin{figure}[h]
	\centering
\subfloat[]{
\includegraphics[width=0.5\linewidth]{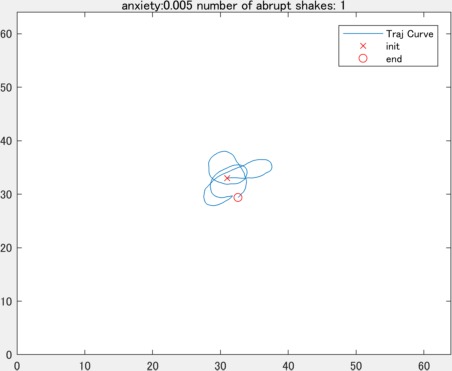}
 }
\subfloat[]{
\includegraphics[width=0.1\linewidth]{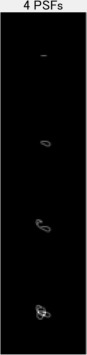}
}

\subfloat[SURF]{
\includegraphics[width=0.23\linewidth]{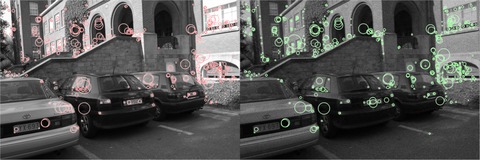}
}
\subfloat[$N_c=189$(SURF)]{
\includegraphics[width=0.23\linewidth]{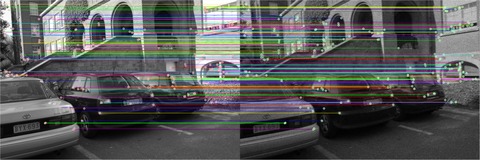}
}
\subfloat[EAS]{
\includegraphics[width=0.23\linewidth]{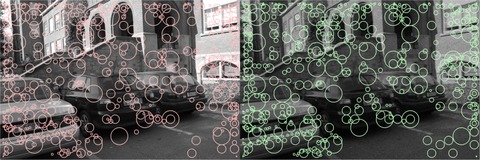}
}
\subfloat[$N_c=340$ (EAS)]{
\includegraphics[width=0.23\linewidth]{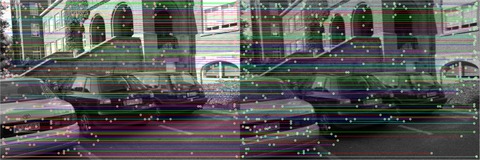}
}

\subfloat[SURF]{
\includegraphics[width=0.23\linewidth]{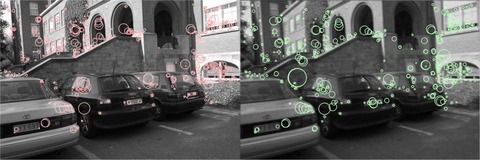}
}
\subfloat[$N_c=3$ (SURF)]{
\includegraphics[width=0.23\linewidth]{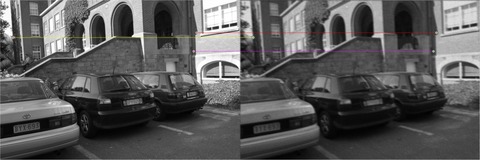}
}
\subfloat[EAS]{
\includegraphics[width=0.23\linewidth]{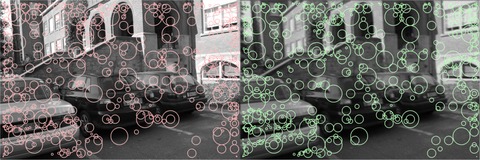}
}
\subfloat[$N_c=246$ (EAS)]{
\includegraphics[width=0.23\linewidth]{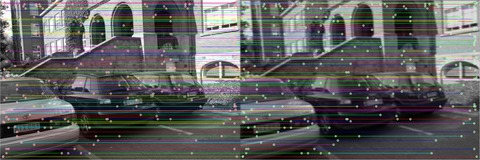}
}

\subfloat[SURF]{
\includegraphics[width=0.23\linewidth]{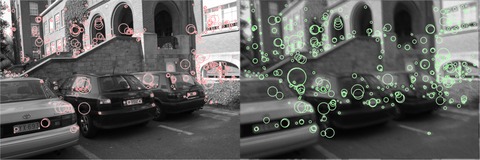}
}
\subfloat[$N_c=22$ (SURF)]{
\includegraphics[width=0.23\linewidth]{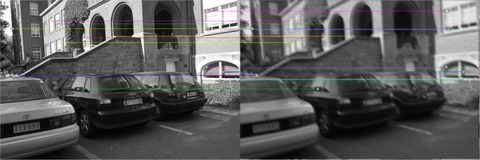}
}
\subfloat[EAS]{
\includegraphics[width=0.23\linewidth]{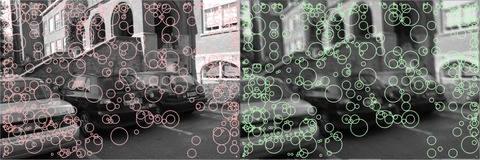}
}
\subfloat[$N_c=265$ (EAS)]{
\includegraphics[width=0.23\linewidth]{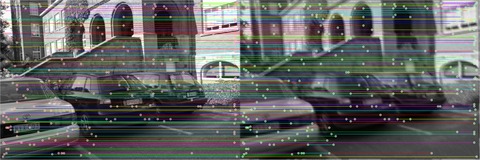}
}

\subfloat[SURF]{
\includegraphics[width=0.23\linewidth]{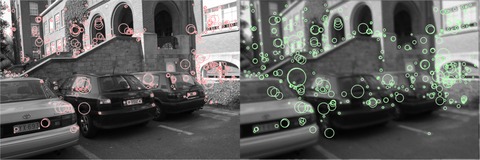}
}
\subfloat[$N_c=20$ (SURF)]{
\includegraphics[width=0.23\linewidth]{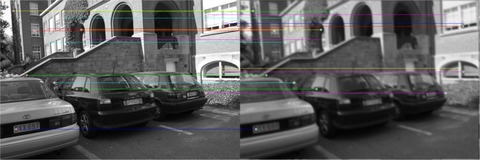}
}
\subfloat[EAS]{
\includegraphics[width=0.23\linewidth]{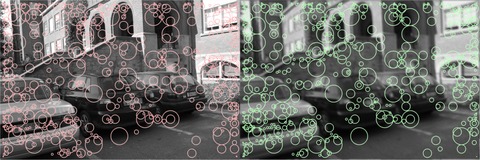}
}
\subfloat[$N_c=289$ (EAS)]{
\includegraphics[width=0.23\linewidth]{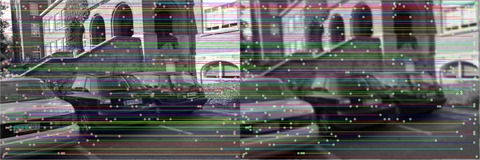}
}
\caption{Keypoint detection results on a Leuven image.}
\label{supp3}
\end{figure}

\begin{figure}[h]
	\centering
\subfloat[]{
\includegraphics[width=0.5\linewidth]{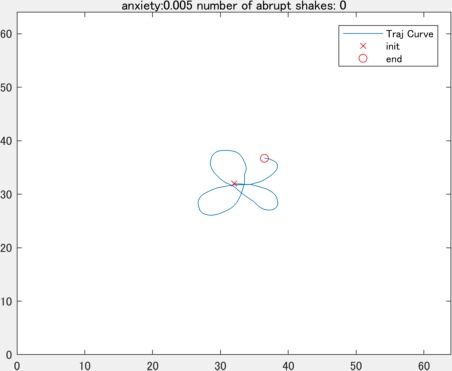}
 }
\subfloat[]{
\includegraphics[width=0.1\linewidth]{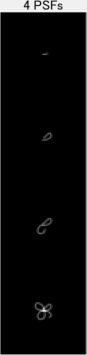}
}

\subfloat[SURF]{
\includegraphics[width=0.23\linewidth]{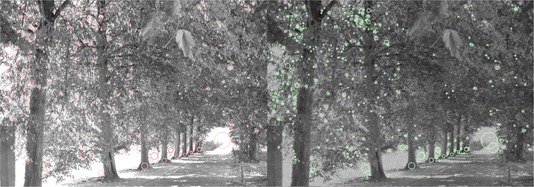}
}
\subfloat[$N_c=15$ (SURF)]{
\includegraphics[width=0.23\linewidth]{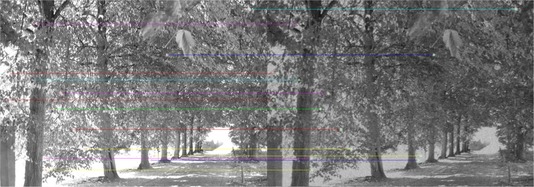}
}
\subfloat[EAS]{
\includegraphics[width=0.23\linewidth]{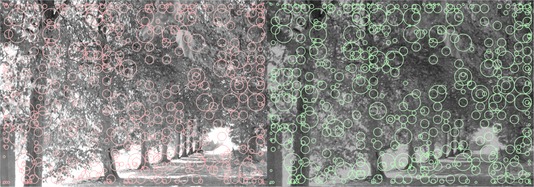}
}
\subfloat[$N_c=335$ (EAS)]{
\includegraphics[width=0.23\linewidth]{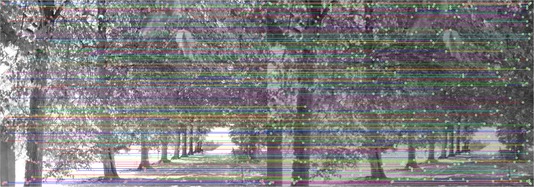}
}

\subfloat[SURF]{
\includegraphics[width=0.23\linewidth]{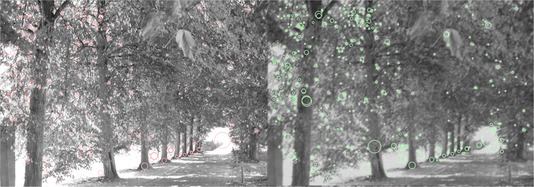}
}
\subfloat[$N_c=1$ (SURF)]{
\includegraphics[width=0.23\linewidth]{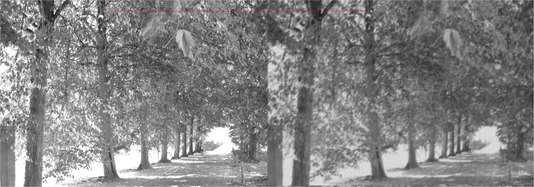}
}
\subfloat[EAS]{
\includegraphics[width=0.23\linewidth]{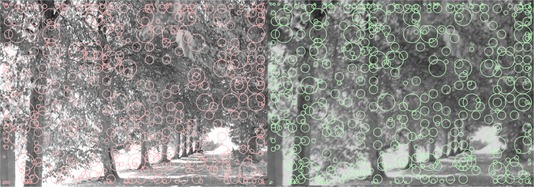}
}
\subfloat[$N_c=250$ (EAS)]{
\includegraphics[width=0.23\linewidth]{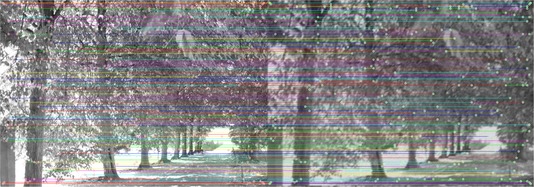}
}

\subfloat[SURF]{
\includegraphics[width=0.23\linewidth]{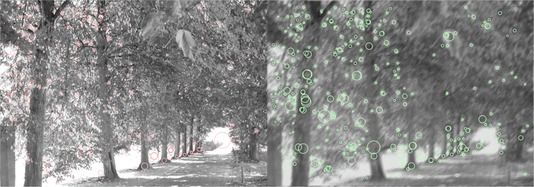}
}
\subfloat[$N_c=25$ (SURF)]{
\includegraphics[width=0.23\linewidth]{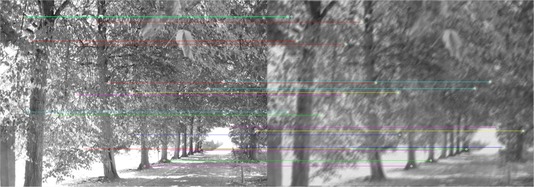}
}
\subfloat[EAS]{
\includegraphics[width=0.23\linewidth]{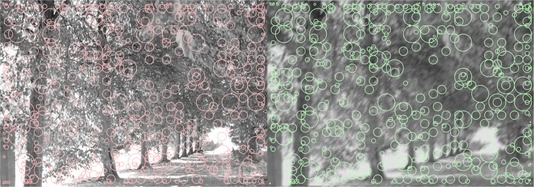}
}
\subfloat[$N_c=318$ (EAS)]{
\includegraphics[width=0.23\linewidth]{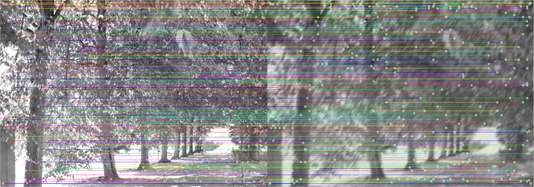}
}

\subfloat[SURF]{
\includegraphics[width=0.23\linewidth]{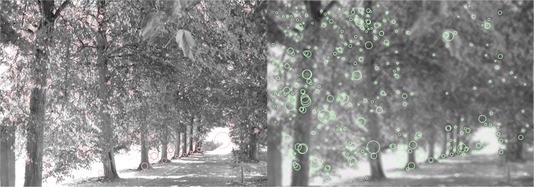}
}
\subfloat[$N_c=8$ (SURF)]{
\includegraphics[width=0.23\linewidth]{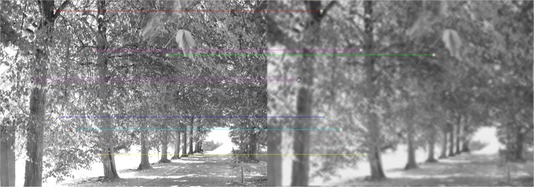}
}
\subfloat[EAS]{
\includegraphics[width=0.23\linewidth]{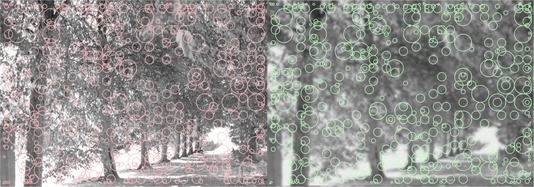}
}
\subfloat[$N_c=306$ (EAS)]{
\includegraphics[width=0.23\linewidth]{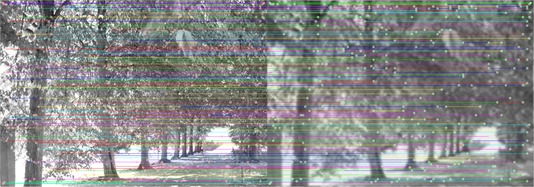}
}
\caption{Keypoint detection results on a Tree image.}
\label{supp4}
\end{figure}

\begin{figure}[h]
	\centering
\subfloat[]{
\includegraphics[width=0.5\linewidth]{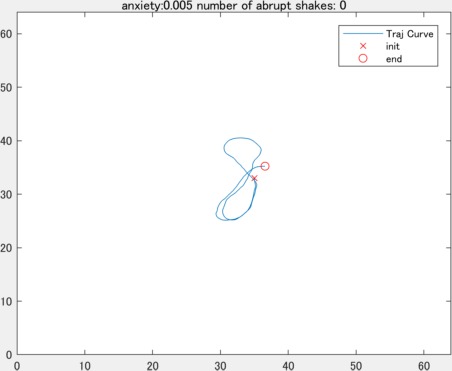}
 }
\subfloat[]{
\includegraphics[width=0.1\linewidth]{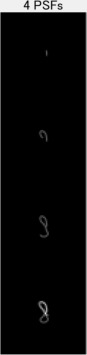}
}

\subfloat[SURF]{
\includegraphics[width=0.23\linewidth]{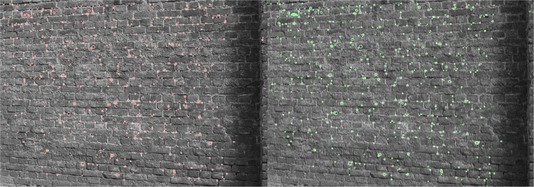}
}
\subfloat[$N_c=2$ (SURF)]{
\includegraphics[width=0.23\linewidth]{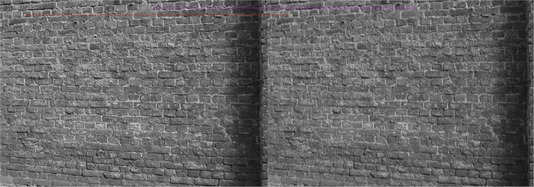}
}
\subfloat[EAS]{
\includegraphics[width=0.23\linewidth]{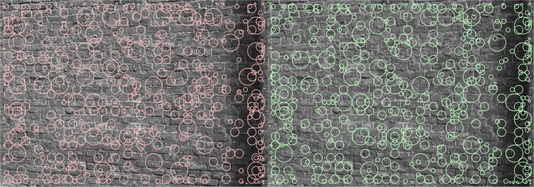}
}
\subfloat[$N_c=262$ (EAS)]{
\includegraphics[width=0.23\linewidth]{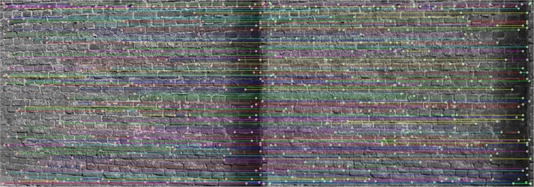}
}

\subfloat[SURF]{
\includegraphics[width=0.23\linewidth]{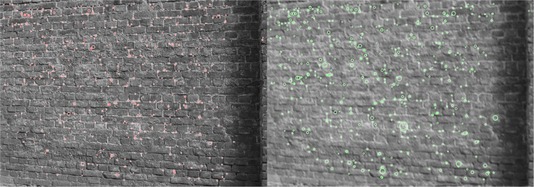}
}
\subfloat[$N_c=2$ (SURF)]{
\includegraphics[width=0.23\linewidth]{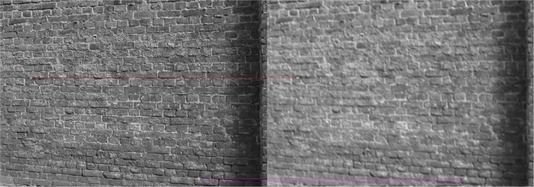}
}
\subfloat[EAS]{
\includegraphics[width=0.23\linewidth]{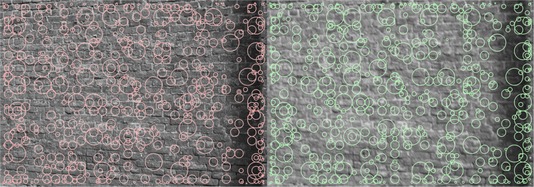}
}
\subfloat[$N_c=206$ (EAS)]{
\includegraphics[width=0.23\linewidth]{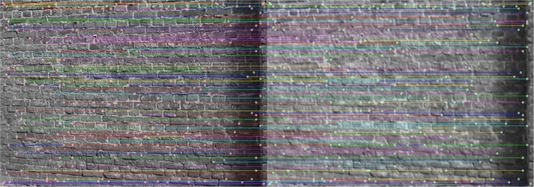}
}

\subfloat[SURF]{
\includegraphics[width=0.23\linewidth]{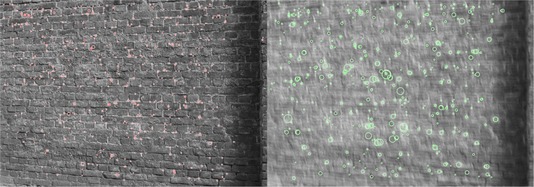}
}
\subfloat[$N_c=2$ (SURF)]{
\includegraphics[width=0.23\linewidth]{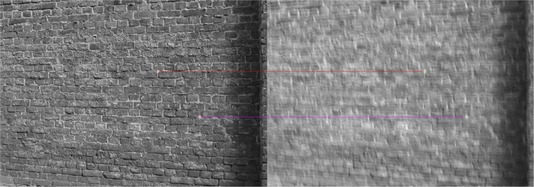}
}
\subfloat[EAS]{
\includegraphics[width=0.23\linewidth]{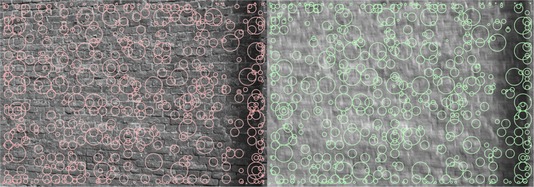}
}
\subfloat[$N_c=282$ (EAS)]{
\includegraphics[width=0.23\linewidth]{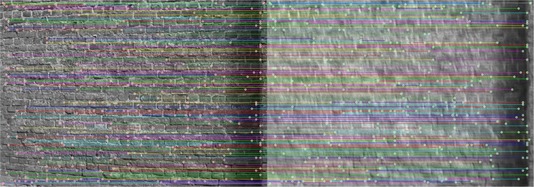}
}

\subfloat[SURF]{
\includegraphics[width=0.23\linewidth]{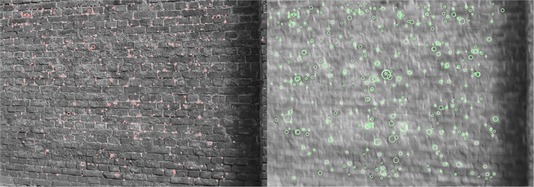}
}
\subfloat[$N_c=4$ (SURF)]{
\includegraphics[width=0.23\linewidth]{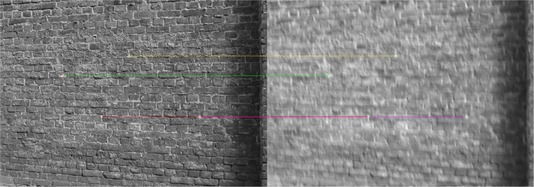}
}
\subfloat[EAS]{
\includegraphics[width=0.23\linewidth]{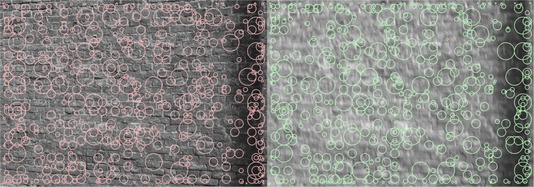}
}
\subfloat[$N_c=285$ (EAS)]{
\includegraphics[width=0.23\linewidth]{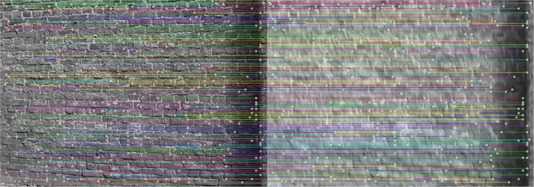}
}
\caption{Keypoint detection results on a Wall image.}
\label{supp5}
\end{figure}

\begin{figure}[h]
	\centering
\subfloat[]{
\includegraphics[width=0.5\linewidth]{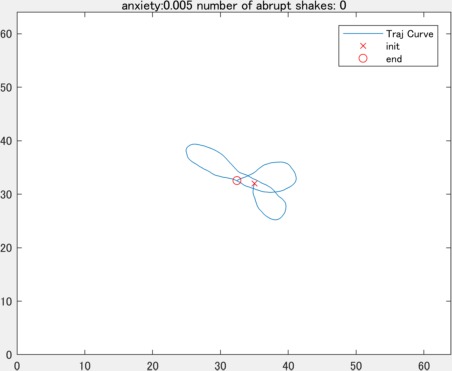}
 }
\subfloat[]{
\includegraphics[width=0.1\linewidth]{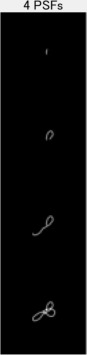}
}

\subfloat[SURF]{
\includegraphics[width=0.23\linewidth]{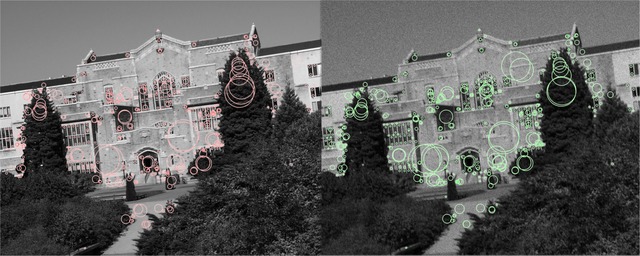}
}
\subfloat[$N_c=2$ (SURF)]{
\includegraphics[width=0.23\linewidth]{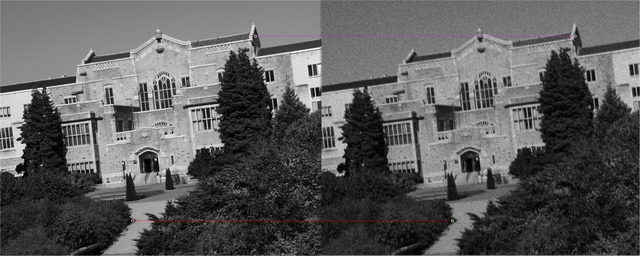}
}
\subfloat[EAS]{
\includegraphics[width=0.23\linewidth]{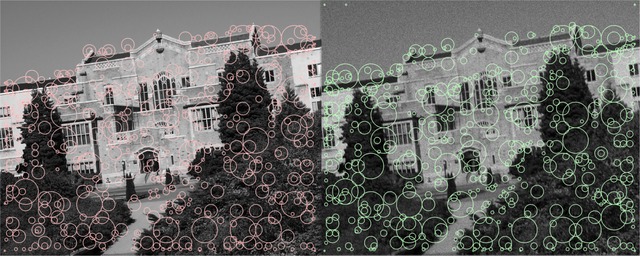}
}
\subfloat[$N_c=213$ (EAS)]{
\includegraphics[width=0.23\linewidth]{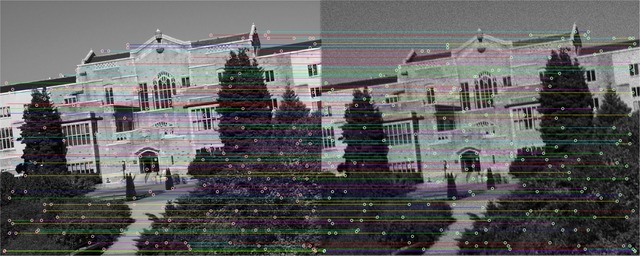}
}

\subfloat[SURF]{
\includegraphics[width=0.23\linewidth]{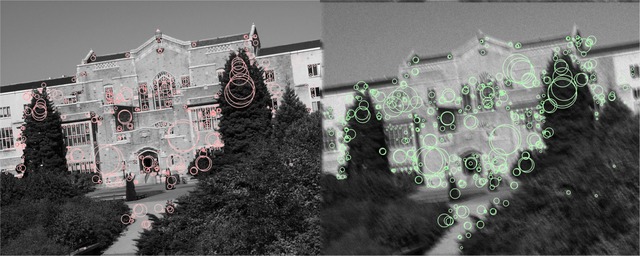}
}
\subfloat[$N_c=2$ (SURF)]{
\includegraphics[width=0.23\linewidth]{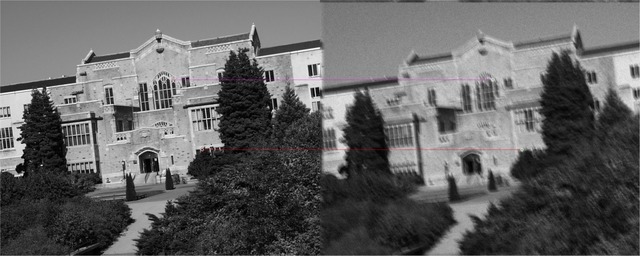}
}
\subfloat[EAS]{
\includegraphics[width=0.23\linewidth]{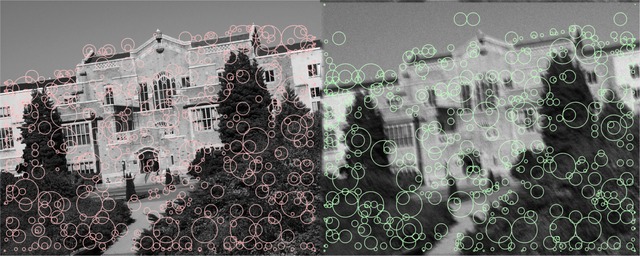}
}
\subfloat[$N_c=94$ (EAS)]{
\includegraphics[width=0.23\linewidth]{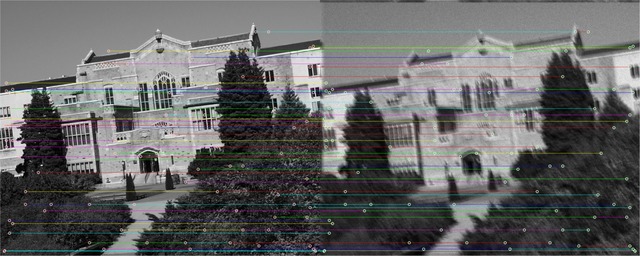}
}

\subfloat[SURF]{
\includegraphics[width=0.23\linewidth]{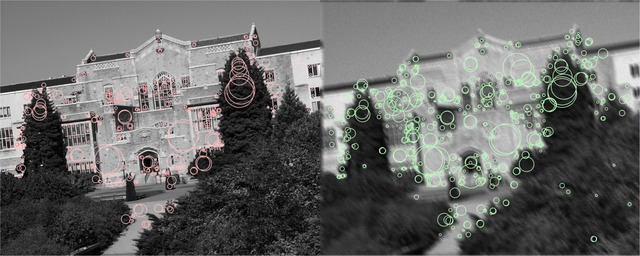}
}
\subfloat[$N_c=1$ (SURF)]{
\includegraphics[width=0.23\linewidth]{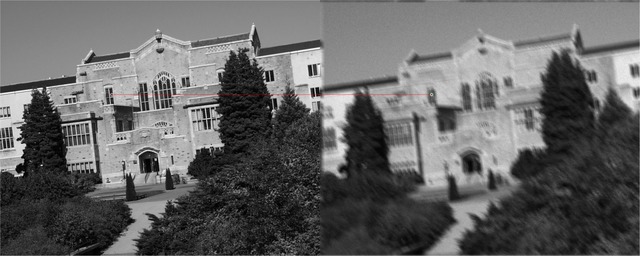}
}
\subfloat[EAS]{
\includegraphics[width=0.23\linewidth]{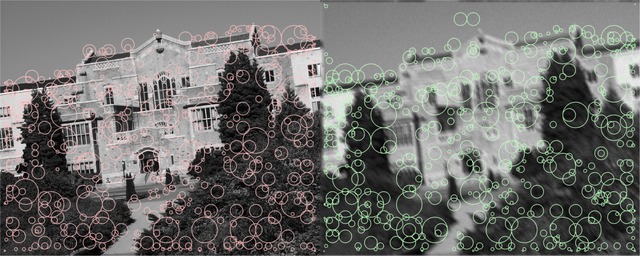}
}
\subfloat[$N_c=124$ (EAS)]{
\includegraphics[width=0.23\linewidth]{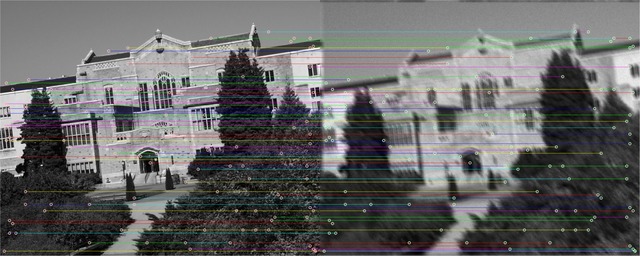}
}

\subfloat[SURF]{
\includegraphics[width=0.23\linewidth]{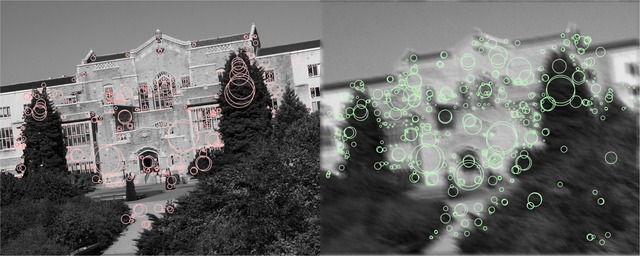}
}
\subfloat[$N_c=2$ (SURF)]{
\includegraphics[width=0.23\linewidth]{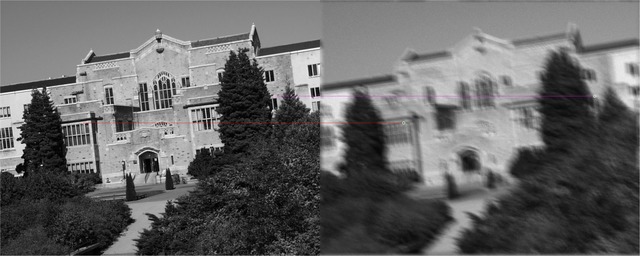}
}
\subfloat[EAS]{
\includegraphics[width=0.23\linewidth]{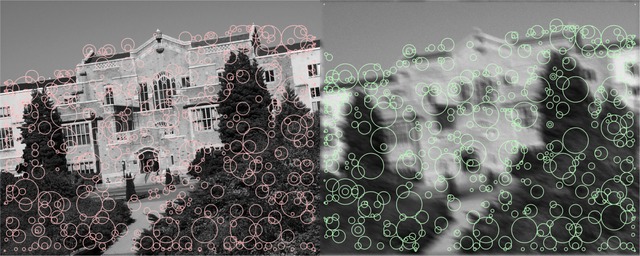}
}
\subfloat[$N_c=160$ (EAS)]{
\includegraphics[width=0.23\linewidth]{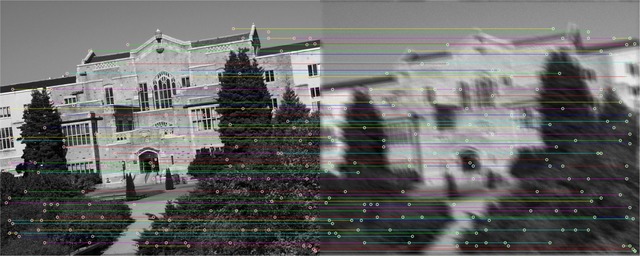}
}
\caption{Keypoint detection results on a UBC image.}
\label{supp6}
\end{figure}

\begin{table}[h]%
 \centering
\caption{Processing time of our method with different image resolutions.}
\begin{tabular}{|l|l|} \hline
  Image resolution & Processing time \\ \hline \hline
  320$\times$240 & 18.80 (ms) \\\hline
  640$\times$480 & 85.76 (ms) \\\hline
  1920$\times$1080 & 530.84 (ms) \\ \hline
  4096$\times$2160 & 2246.74 (ms) \\\hline
 \end{tabular}
 \label{table2}%
\end{table}
To further demonstrate the robustness of our method, we show results under relatively more realistic blur. Specifically, we adopt the method described in \cite{boracchi2012modeling} to generate general motion blur with random motion trajectories, PSFs and sensor noise depending on the exposure time. From Fig. \ref{supp1} to Fig. \ref{supp6}, (a) shows the random motion trajectory, and (b) shows four types of PSFs generated from the trajectory. Exposure time from the top row to bottom row in each figure is set to 0.0625s, 0.25s, 0.5s and 1s, respectively. We show comparative results using fast Hessian from SURF\cite{bay2006surf} and our EAS. $N_c$ denotes the number of pairs of corresponding keypoints. As we can observe from the results, EAS outperforms fast Hessian in all the cases.

\subsection{Quantitative Results}
\begin{figure}[b]
	\centering
  
\subfloat{
\includegraphics[width=0.75\linewidth]{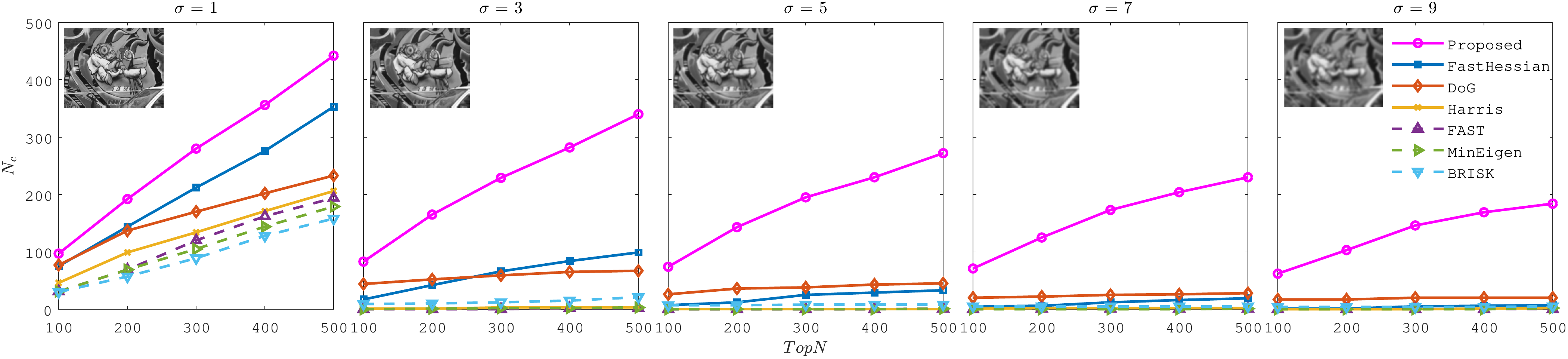}
 \label{fig:Graffiti_Gaussian}
 }
 
\subfloat{
\includegraphics[width=0.75\linewidth]{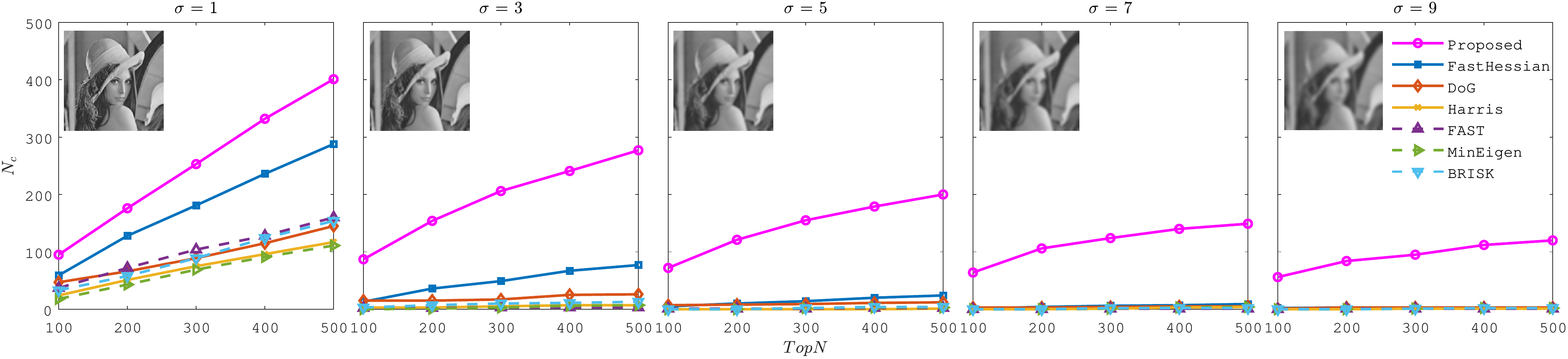}
 \label{fig:Lena_Gaussian}
}

\subfloat{
\includegraphics[width=0.75\linewidth]{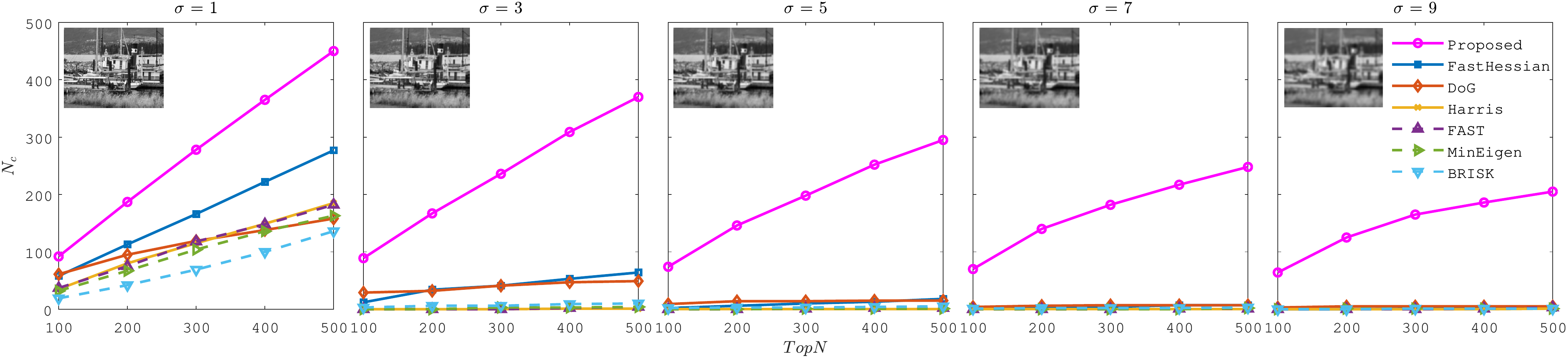}
 \label{fig:Boat_Gaussian}
}
\caption{Quantitative results ($N_c$ v.s. $TopN$) on images corrupted with Gaussian blur. Blur becomes severer from left to right. First to third row are respectively Grafitti, Lena and Boat. }
\label{fig:comparison_Gaussian}
\end{figure}

\begin{figure}[b]
	\centering
\subfloat{
\includegraphics[width=0.75\linewidth]{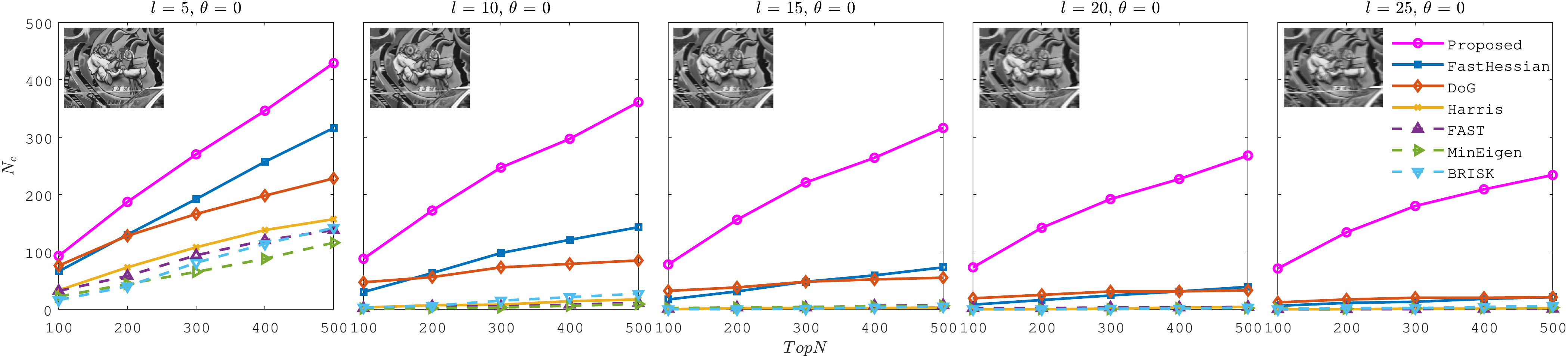}
 \label{fig:Graffiti_Motion}
}

\subfloat{
\includegraphics[width=0.75\linewidth]{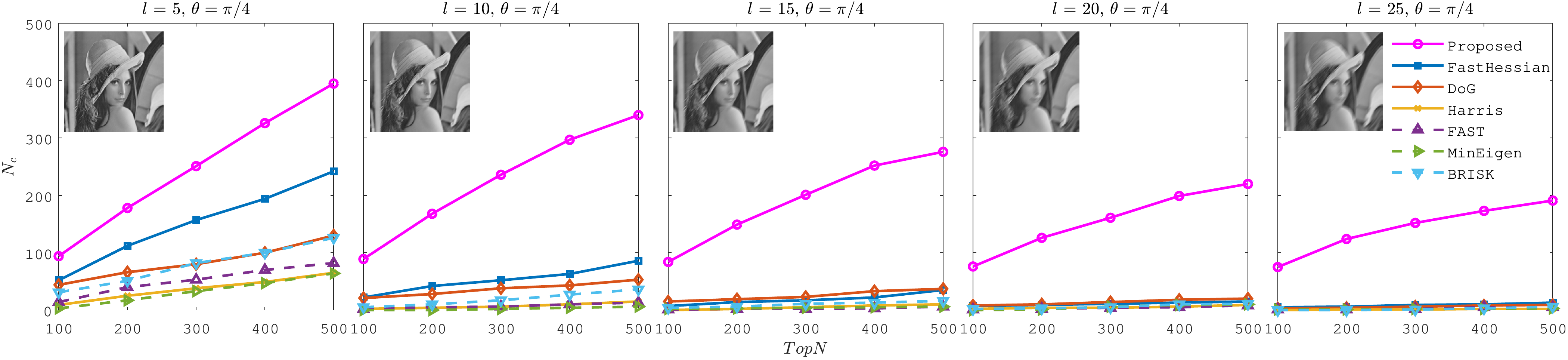}
 \label{fig:Lena_Motion}
}

\subfloat{
\includegraphics[width=0.75\linewidth]{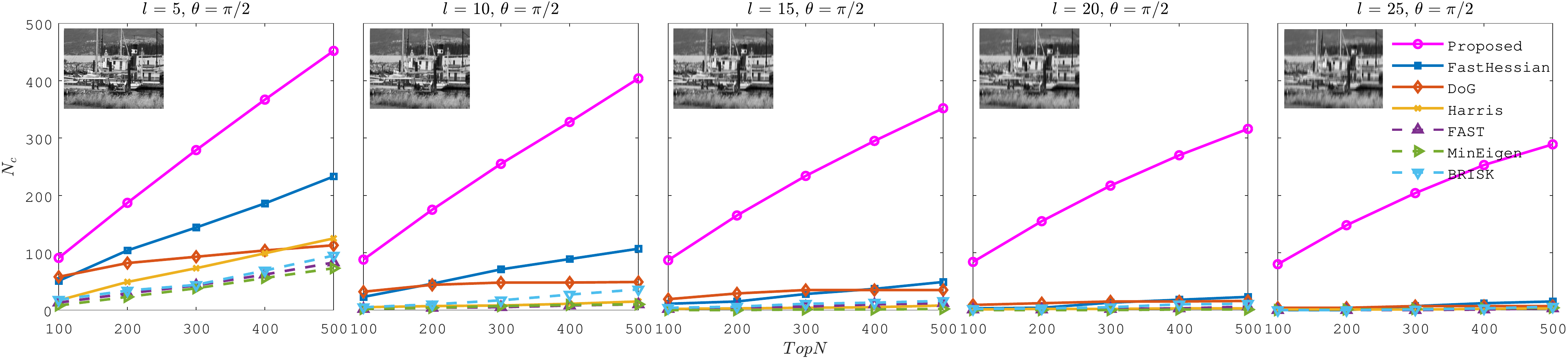}
 \label{fig:Boat_Motion}
}
\caption{Quantitative results on images with motion blur. Blur becomes severer from left to right. First to third row are respectively Grafitti, Lena and Boat.}
\label{fig:comparison_motion}
\end{figure}

Fig. \ref{fig:comparison_Gaussian} and Fig. \ref{fig:comparison_motion} show the quantitative results with respect to Gaussian blur and linear motion blur, with the $x$ axis representing $TopN$ and $y$ axis representing $N_c$. Specifically, the results over ``3 images $\times$ 2 types of blur $\times$ 5 degrees of blur $\times$ 5 $TopN$ $\times$ 7 methods'' are studied. The overall trend observed is that with the increase of $TopN$, $N_c$ increases, unless the degree of blur exceeds the ability of each method. Also, for all the methods, $N_c$ decreases with the increase of blur degree. EAS is observed to be robust against the increase of blur degree. Another interesting overall finding is that in the case of slightly blurred images, $N_c$ of EAS increases linearly along with the increase of $TopN$. In general, the curve is likely to converge with the increase of $TopN$ with heavy blur.

In Fig. \ref{fig:comparison_Gaussian}, the parameter $\sigma$ of Gaussian varies from 1 to 9. For all the images, conventional methods can only detect few valid keypoints when $\sigma$ is greater than 5. Although the curve of EAS converges to a certain limit and moves to lower right with respect to the increase of $\sigma$, our EAS achieves 37.2\% of average repeatability which is much higher than Fast-Hessian (10.4\%). In Fig. \ref{fig:comparison_motion}, the parameter $l$ of linear motion blur varies from 5 to 25, and $\theta=0, \pi/4, \pi/2$ are respectively applied to ``Graffiti'', ``Lena'', ``Boat''. As a final result, 42.3\% of average repeatability can be obtained by EAS, which soundly outperforms Fast-Hessian (11.6\%). The examples of processing time shown in Table \ref{table2} are measured with complete MATLAB implementation (without parallelization or optimization) on a i7-7700 CPU@3.60GHz, 32.0GB RAM desktop computer.




\section{Conclusions}
In this paper, we presented a metric for measuring the distance between two derivative distributions, based on the difference of the sum of eigenvalues. Furthermore, EAS for measuring asymmetry is proposed for keypoint detection. It is robust in detecting corresponding points under various types and degrees of blur. Extensive experiments demonstrate that our EAS outperforms the state of the art methods in the presence of image blur. 

Despite the robustness of our method, it still has a few limitations. It is likely to detect more keypoints in unblurred regions when both unblurred and blurred regions exist. One potential way to solve this problem is using supervised signals (e.g., classification for blurred/unblurred regions). Intensity variations around the detected corresponding points could be small, thus leading to difficulties for the description of the gradient based local feature descriptors. As the future work, we would like to develop effective scale search methods for EAS and design blur-countering feature descriptors for real-world applications.

\clearpage
\bibliographystyle{elsarticle-num}
\bibliography{egbib}

\end{document}